\documentclass{article} % For LaTeX2e
\usepackage{natbib} 
\usepackage{times}
\usepackage{amsmath,amsfonts,bm}

\def\eqref#1{equation~\ref{#1}}
\def\1{\bm{1}}

\DeclareMathAlphabet{\mathsfit}{\encodingdefault}{\sfdefault}{m}{sl}
\SetMathAlphabet{\mathsfit}{bold}{\encodingdefault}{\sfdefault}{bx}{n}

\usepackage{hyperref}
\usepackage{url}
\usepackage{graphicx}
\usepackage{subcaption}
\usepackage{algorithm}
\usepackage{algorithmic}
\usepackage[title]{appendix}
\usepackage[
    left=1.5in,      % Sets the left margin
    right=1.5in,     % Sets the right margin
    top=1.85cm,         % Sets the top margin (influences header position)
    headheight=10pt, % Height of the header box
    headsep=8pt     % Separation between the header and the main text
]{geometry}

\title{Function regression using the forward forward training and inferring paradigm}

\author{
  Shivam Padmani\\
  Department of Mechanical Engineering\\
  Indian Institute of Science \\
  Bangalore \\
  \and
  Akshay Joshi\thanks{Corresponding author: \texttt{akshayjoshi@iisc.ac.in}}\\
  Department of Mechanical Engineering\\
  Indian Institute of Science \\
  Bangalore
}
\begin{document}
\maketitle
\begin{abstract}
Function regression/approximation is a fundamental application of machine learning. Neural networks (NNs) can be easily trained for function regression using a sufficient number of neurons and epochs. The forward-forward learning algorithm is a novel approach for training neural networks without backpropagation, and is well suited for implementation in neuromorphic computing and physical analogs for neural networks. To the best of the authors' knowledge, the Forward Forward paradigm of training and inferencing NNs is currently only restricted to classification tasks. This paper introduces a new methodology for approximating functions (function regression) using the Forward-Forward algorithm. Furthermore, the paper evaluates the developed methodology on univariate and multivariate functions, and provides preliminary studies of extending the proposed Forward-Forward regression to Kolmogorov Arnold Networks, and Deep Physical Neural Networks.
\end{abstract}

\section{Introduction}
The computational demands associated with the training and inference of AI models result in substantial energy consumption. As the adoption of AI continues to grow exponentially, the associated escalating energy requirement also poses significant challenges, necessitating the development of energy-efficient alternatives for computational processes \citep{mehonic2022brain}.
Brain-inspired (Neuromorphic) computing paradigms are designed to mimic the brain's computing processes, in order to translate the energy efficiency of the neural connections in the brain to computing tasks. Other analog or physical systems can also be considered useful tools for developing energy-efficient solutions for computing \citep{zolfagharinejad2024brain}. An example of a physical computing device would be the memristor called ``dot-product engine" \citep{li2023memristive,zhang2020brain,chen2023electrochemical}. It serves as a physical analog for matrix-vector multiplication and performs the multiplication in a single step instead of the usual $n^2$ steps in traditional computing methods \citep{sharma2024linear}.

Neural Networks (NNs) are fundamental building blocks to deep learning frameworks. Their implementation on classical digital computers is energy-intensive. To address this high energy consumption, recent trends have focused on exploring physical systems that may serve as analogs to digital neural networks \citep{wright2022deep}. These are popularly called physical neural networks. Physical neural networks use physical systems or materials to emulate the behavior of neurons and synapses. Neural networks use complex activation functions to add non-linearity to the system. Similarly, these physical systems use physical phenomena that help surrogate these activation functions and layer-wise computations. Recent works have demonstrated that using physical systems for neural network computations is not only highly energy-efficient but also achieves above 90\% classification accuracy \citep{wright2022deep,momeni2023backpropagation}.

Training of neural networks is popularly done using the backpropagation algorithm \citep{linnainmaa1970representation,linnainmaa1976taylor,griewank2012invented}.  The backpropagation (BP) algorithm uses supervised learning to optimize the loss function using methods like stochastic gradient descent. However, the BP algorithm is highly energy inefficient because of the need for forward and backward passes for each step of the optimization. The backward pass calculates the gradient of the loss function with respect to each parameter in the network. This requires a lot of energy and time, which only increases with increase in depth and complexity of the NN architecture. In addition, backpropagation-based parameter learning is not suitable for multi-layered physical neural networks which are unidirectional in time. Earlier works implementing physical neural networks \citep{wright2022deep} used digital twins of the physical system to achieve backpropagation during training. However, digital twins add to the energy cost during the training process and are not accurately available for many physical systems. Therefore, the ability to train NNs without BP has the potential to significantly improve the energy efficiency of the aforementioned physical NNs.

In 2022, \citet{hinton2022forward} proposed a new algorithm called the Forward-Forward (FF) algorithm, which uses only a forward pass to train the neural networks. This neuromorphic algorithm is based on the idea of learning by comparing correctly labeled and incorrectly labeled data. As the FF algorithm is unidirectional, it can be used to train physical neural networks. Layer-wise training in this algorithm trains each layer of the network to correctly distinguish between correctly labeled and incorrectly labeled data points. Subsequent to the introduction of the FF algorithm, various researchers have extended it to various applications like convolutional neural networks \citep{scodellaro2023training}, recurrent neural networks \citep{kag2021training}, etc. Some have also developed variations of this algorithm to enhance it for better accuracy \citep{wu2024distance,lorberbom2024layer}. Some researchers have also successfully implemented the FF approach to train neural networks using physical systems combined with digital linear layers \citep{momeni2023backpropagation}. The FF Algorithm is designed to solve classification problems and to the best of the authors' knowledge all of the prior works only address classification tasks. In this work, we seek to extend the FF algorithm for the function regression task. In section \ref{sec:Theory} we discuss the theoretical background of the FF algorithm and how it can be extended to regression tasks, with section \ref{sec:Results} discussing the 1-D, 2-D and 3-D benchmarks for validating the proposed FF-regression algorithm and section \ref{sec:conc} concluding the manuscript.

\section{Theoretical Background}
\label{sec:Theory}
% \subsection{Forward Forward Regression}
\subsection{Forward Forward training}
\begin{figure}[h]
    \centering
    \begin{subfigure}[b]{0.4\textwidth}
        \includegraphics[width=\linewidth]{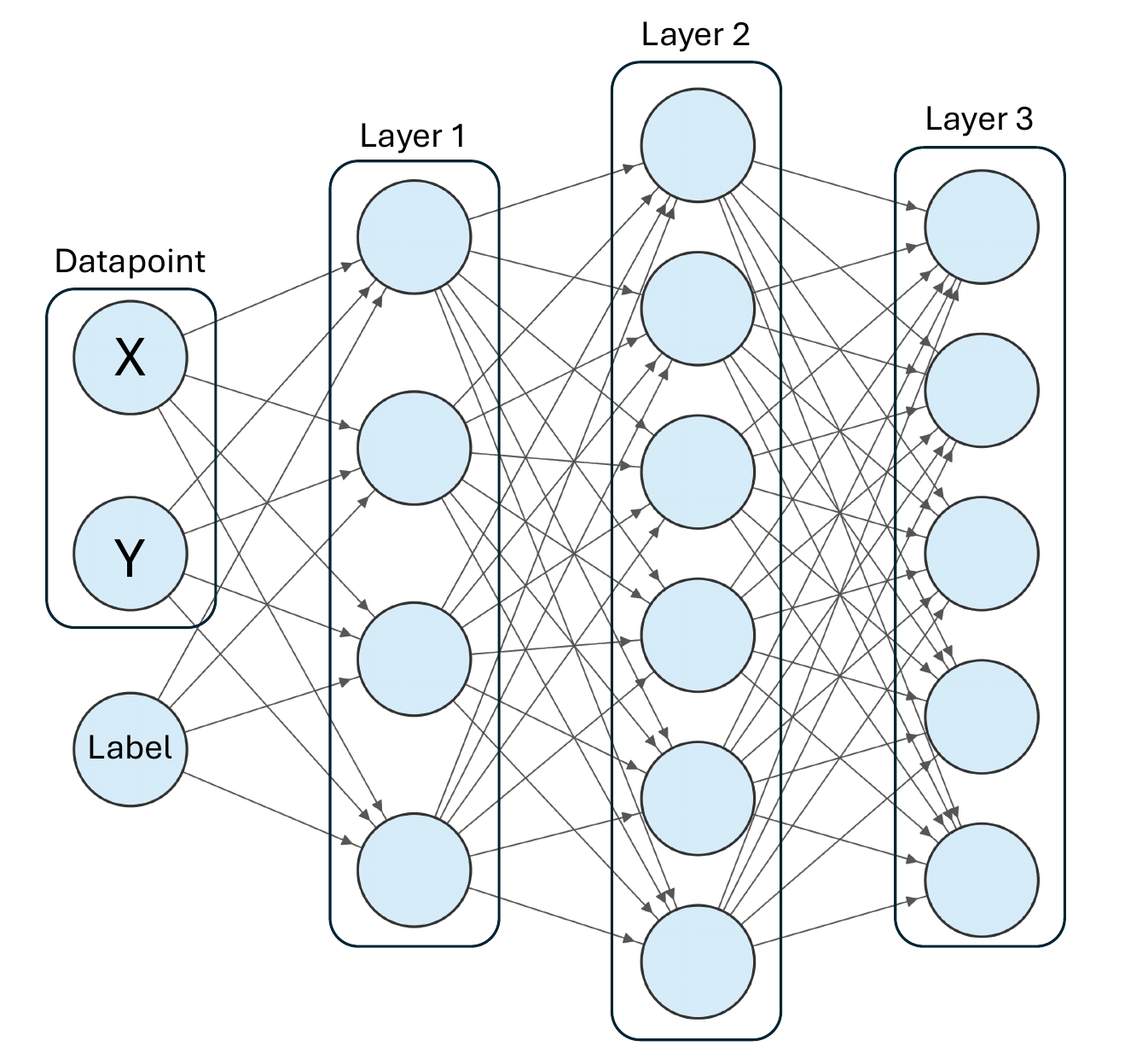}
        \caption{}
        \label{fig:ffnn}
    \end{subfigure}
    \hspace{1.0cm}
    \begin{subfigure}[b]{0.4\textwidth}
        \includegraphics[width=\linewidth]{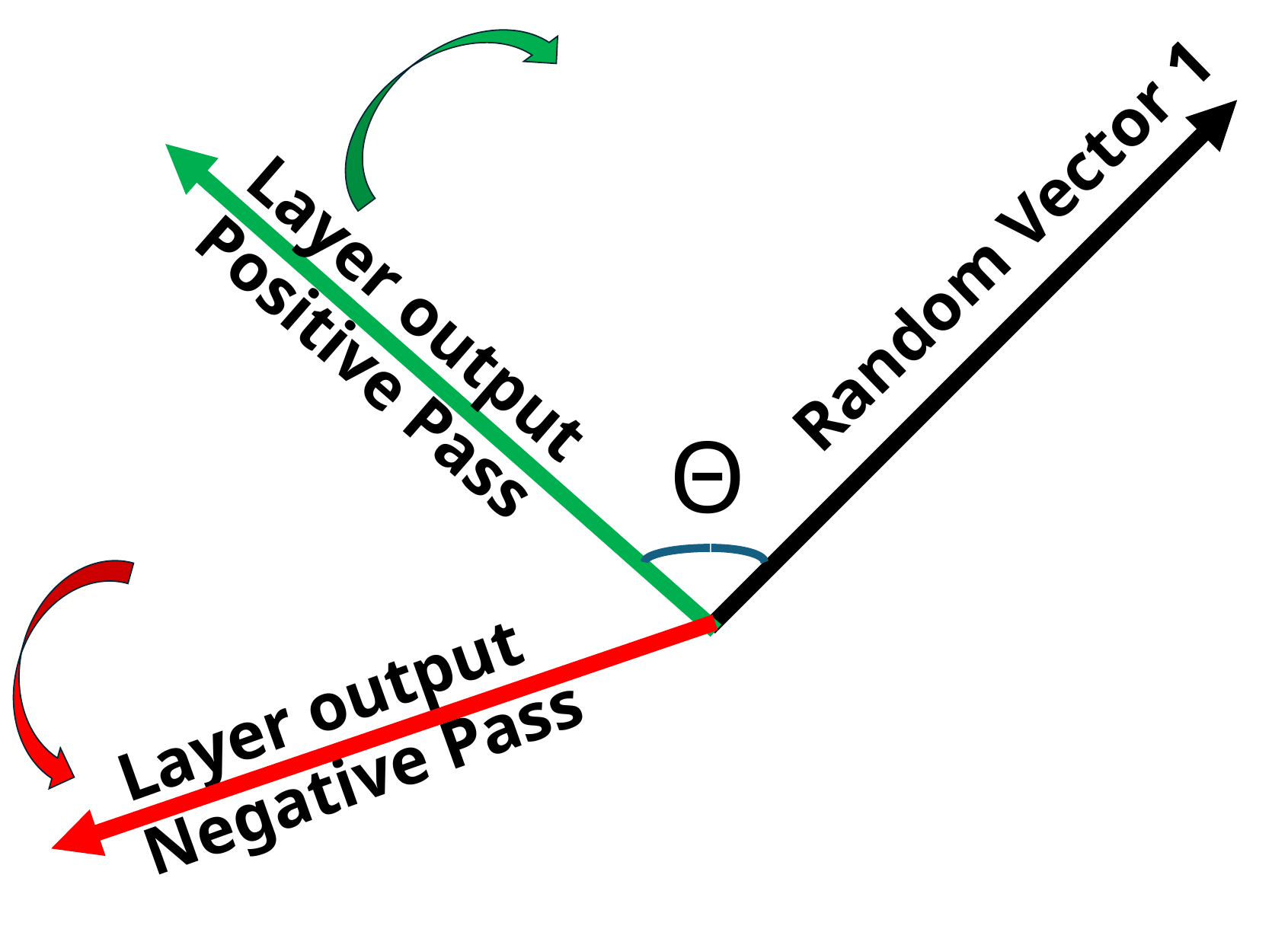}
        \caption{}
        \label{fig:goodness_vector}
    \end{subfigure}
    \caption{(a) Schematic diagram of neural network which can be trained using forward forward algorithm. Note that the last layer is also the same as a hidden layer, i.e., there is no output from the final layer. (b) Arbitrary vectors used to optimize the difference between positive and negative goodness.}
    \label{fig:fig1}
\end{figure}

The Forward-Forward algorithm is an approach to train neural networks layer-wise without using backpropagation, by relying on learning by comparison. This is developed from the idea of contrastive learning, where both correct and incorrect data are required for training the model by comparison \citep{khosla2020supervised}. This algorithm takes two types of data-- correctly labeled (positive) and incorrectly labeled (negative) data. A function called ``goodness'' quantifies each layer's output. The input for the goodness function of a layer is the layer's output. The output of the layer's goodness function is a scalar value. The goodness value for positive and negative datasets is calculated separately. The weights of the layer are optimized to maximize the difference between the goodness for the positive data and the goodness for the negative data. Thus, each layer in the network is optimized to discriminate between positive (correctly labeled) and negative (incorrectly labelled) data. Furthermore, this is done serially (layer-by-layer) and independently without any backward propagation of gradients. A schematic of a NN trained and inferred using the FF paradigm is provided in Fig. \ref{fig:ffnn}. 

\subsection{Discussion on Goodness function}\label{sec:discgoodness}
The goodness function can be any function that receives the outputs of a NN layer as an argument and returns a scalar value. A higher value of goodness for a layer indicates that the datapoint-label combination are correct (correctly labeled), while a lower value of goodness indicates that the combination is incorrect (incorrectly labeled). In \cite{hinton2022forward}, it was suggested to define goodness function as the sum of squares of layer output. The outputs of a layer are then normalized before being passed as inputs to the next layer in order to avoid biasing subsequent layers. Another more recent work \citep{momeni2023backpropagation} used cosine similarity as a goodness function. The cosine similarity is evaluated between a layer output and a fixed random vector of equal dimension. As illustrated in Fig. \ref{fig:goodness_vector}, the objective for training the layer is to increase the cosine similarity between the layer output and the vector for positive data and to reduce it for negative data. This training increases the gap between goodness associated with positive data (correctly labeled) and negative data (incorrectly labeled). Note that arbitrary vectors should be different for different layers of the neural network. The particular definition of the goodness function adopted in \cite{momeni2023backpropagation} has an added advantage of not requiring renormalization of layer outputs at every layer. In the current work, the goodness function used in \cite{momeni2023backpropagation} is adopted.

\subsection{Discussion on loss function and training}

During training, the dataset is categorized into positive (correctly labeled) and negative (incorrectly labeled). The goodness associated with the positive and negative data for the $i^{th}$ layer is written as in Eqs. \ref{eqn:cossimpositive} and \ref{eqn:cossimnegative}:
\begin{equation}
g_{\text{pos}}^{(i)} = cos_{sim}(y_{\theta_{(i)},\text{pos}}^{(i)}, \zeta^{(i)}) \label{eqn:cossimpositive}
\end{equation}
\begin{equation}
g_{\text{neg}}^{(i)} = cos_{sim}(y_{\theta_{(i)},\text{neg}}^{(i)}, \zeta^{(i)}) \label{eqn:cossimnegative}
\end{equation}
where, $g_{\text{pos}}^{(i)}, g_{\text{neg}}^{(i)} \in {\rm I\!R} $ are the goodness values of positive data and negative data, respectively, associated with the outputs $y_{\text{pos},\theta_{(i)}}^{(i)}, y_{\text{neg},\theta_{(i)}}^{(i)} \in {\rm I\!R}^{d_{i}} $ of layer $i$.
$\zeta^{(i)} \in {\rm I\!R}^{d_{i}}$ is a fixed arbitrary vector of dimension $d_{i}$. The goodnesses are evaluated as the cosine similarity ($cos_{sim}$) between the vectors $y^{(i)}_{\theta_{(i)}}$ and $\zeta^{(i)}$.

Given that the training is performed layer-wise, the Loss function is defined for each layer. As discussed in section \ref{sec:discgoodness}, the minimization of the layer's loss function should lead to the maximization of the difference between $g_{\text{pos}}$ and $g_{\text{neg}}$. Following \cite{momeni2023phyff} the layer-wise loss function used in the current work is given in Eq. \ref{eqn:Loss_i}
\begin{equation}
Loss^{(i)} = \log(1+\exp(-\theta \delta^{(i)}))\label{eqn:Loss_i}
\end{equation}
where $\delta^{(i)} = g_{\text{pos}}^{(i)}-g_{\text{neg}}^{(i)}$
From Eq. \ref{eqn:Loss_i} it is evident that the layer's loss is minimized by maximizing $\delta^{(i)}$, i.e, by maximizing $g_{\text{pos}}$ and minimizing $g_{\text{neg}}$.

\subsection{Inferencing in Forward Forward Neural Networks}
\label{subsec:inferFFNN}
Regular NNs trained using the BP paradigm have neural connections that flow between the input neuron and the output neuron(s). However, as seen in Fig. \ref{fig:ffnn}, neural networks using the FF paradigm do not have such input-to-output neural connections. Instead, the ``input" data point X and the ``output" data point Y both feature in the input neural layer alongside the label L, i.e, the classification label for the data point is encoded as a part of the input to the neural network. As discussed in the previous sections, each layer of the Forward-Forward NNs are trained to provide low goodness outputs in cases where the datapoint-label combination is incorrect, and provide high goodness outputs for correct datapoint-label combinations. During forward inferencing, the label which provides the highest sum of goodness outputs across all layers is selected as the ``correct" label corresponding to the data point. This implies that the forward pass must be performed for all labels for a given data point in order to identify the correct label/classification for the given data. In comparison, NNs trained using BP would require a single forward pass during inferencing.

\subsection{Function Regression using Forward Forward approach}\label{subsec:FFregression}

\begin{figure}[!ht]
    \centering
    \includegraphics[width=0.8 \linewidth]{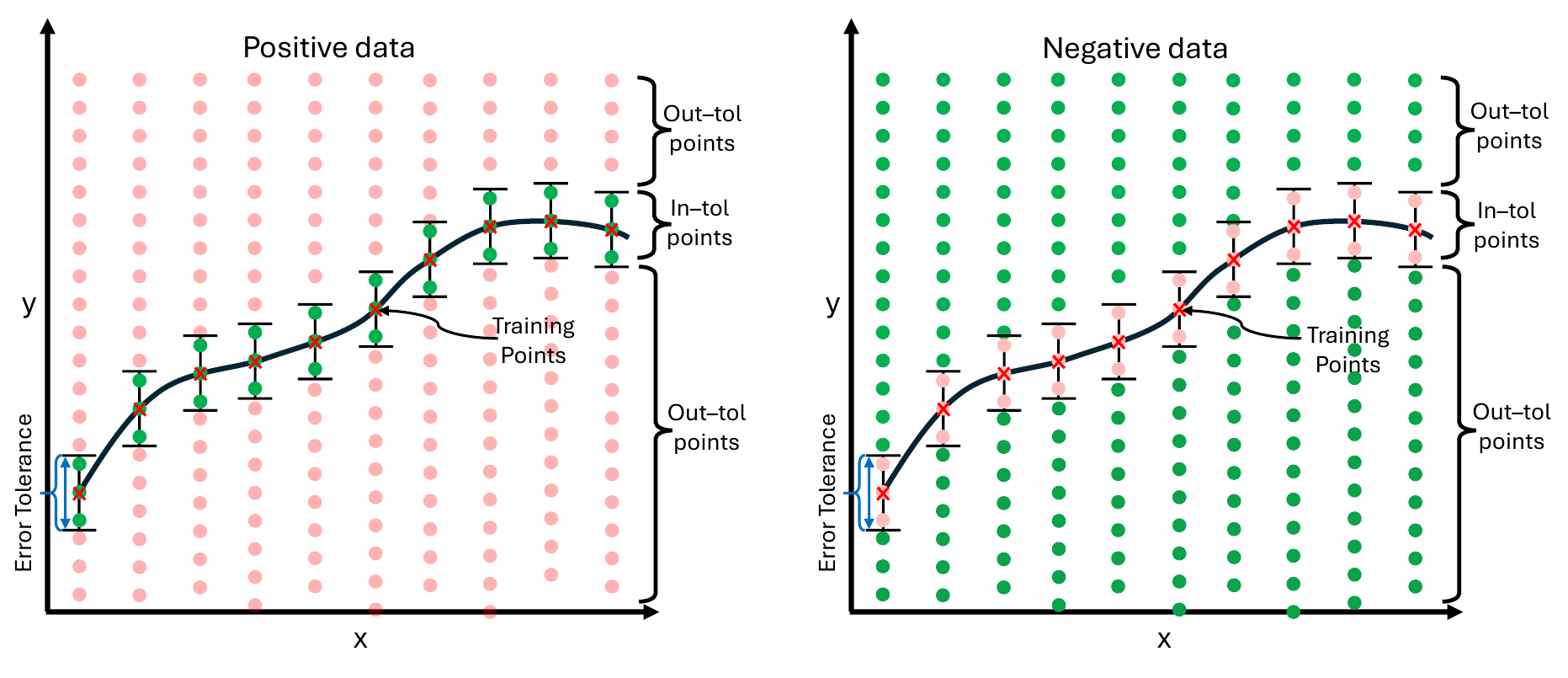}
    \caption{Schematic diagram for training of a FF NN, with the red crosses indicating the training data-points, and colored circles indicating the trial points with green corresponding to label 1.0 and red corresponding to label 0.0. On the left, the positive data has the trial points labeled correctly, while the right figure shows the negative data with incorrect labeling.}
    \label{fig:trainingdata}
\end{figure}

% Figure 1 description
As discussed in the preceding sections, the FF algorithm is primarily suited to classification tasks wherein the input datapoint-label combination is classified as either ``correct" or ``incorrect" by the NN. In this context, we can view the task of function regression as classification of datapoints as either within a preset tolerance level of the training data, or outside of the preset tolerance level of the training data. The case of training a FF NN to approximate a 1-D function is illustrated in figure \ref{fig:trainingdata}, where the training points are highlighted as blue crosses, and the trial points are the colored circles. The errorbars highlight the user-defined tolerance level (tol), within which a trial point is considered to belong to the function-value (in-tol). All trial points outside the errorbar are considered to be not equal to the function-value (out-tol). All in-tol points are colored in green and out-tol points are colored in red for visual illustration. Next, we chose a labelling scheme that assigns the label 1 to in-tol points and 0 to out-tol points. Thus, to train a FF NN, the positive (correctly labeled) dataset will consist of 1 assigned to in-tol points and 0 to out-tol points and the negative dataset will have 1 assigned to out-tol points and 0 assigned to in-tol points. As shown in figure \ref{fig:ffnn}, the inputs to training a FF NN would be the x and y co-ordinate of the trial points and the associated labels. Each layer in the FF NN would be trained to minimize Eq. \ref{eqn:Loss_i}, i.e., maximize the difference in goodness output between the positive and negative dataset. Thus, a well-trained FF NN is expected to classify any given point in space as either in-tol and out-tol. Algorithm \autoref{alg:training} summarizes the training of a FF NN for regression.

\begin{figure}[!ht]
    \centering
    \includegraphics[width=0.6\linewidth]{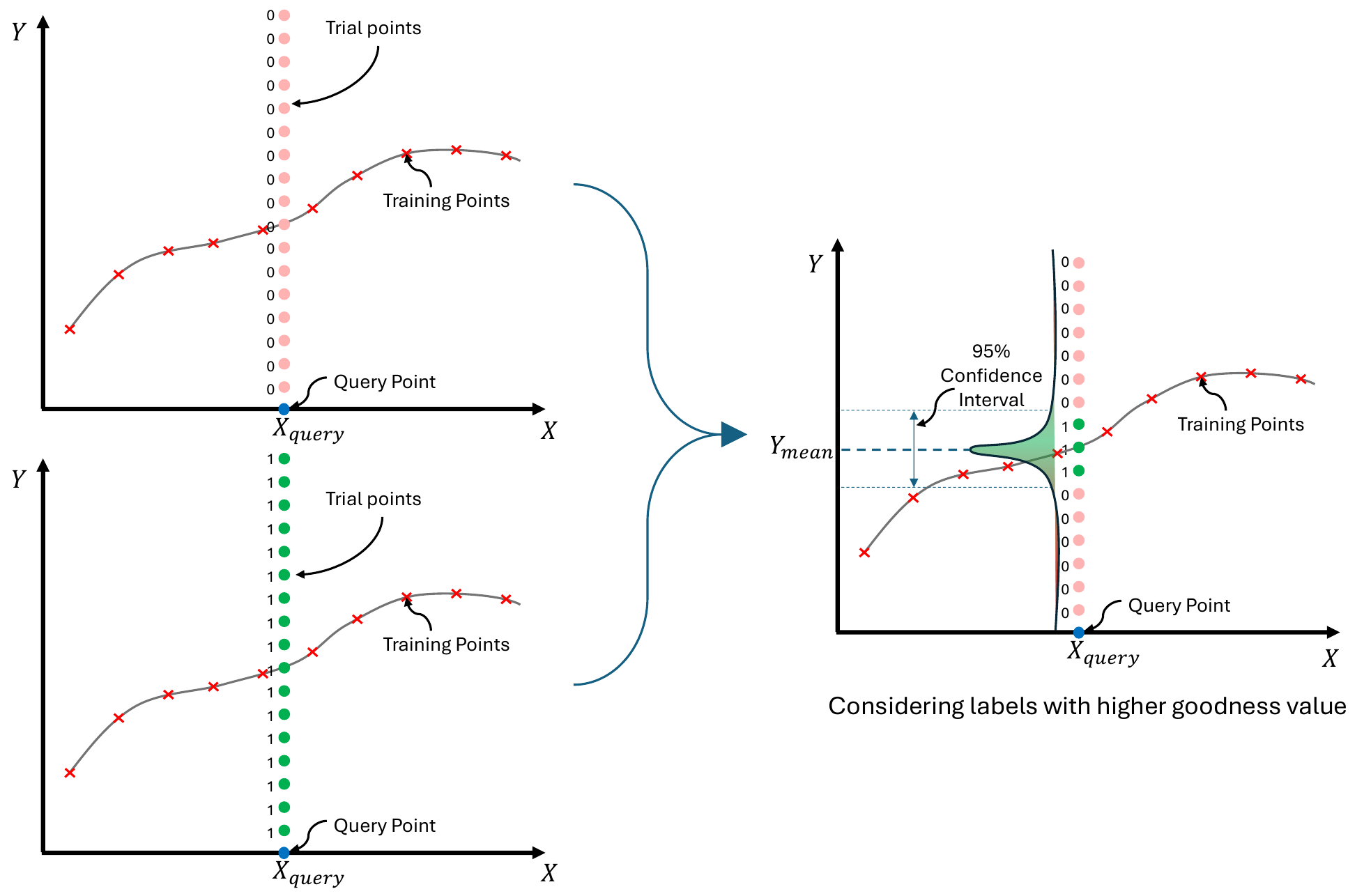}
    \caption{Schematic diagram of prediction phase while inferencing from the trained forward forward neural network. Both labels-- in-tol (1.0) and out-tol (0.0) are applied to all the trial points. The label yielding the higher goodness value would ideally be chosen as the correct label for the trial point (see \autoref{subsubsec:goodnessinvert}).}
    \label{fig:predictionphase}
\end{figure}

We make use of this ability of the FF NN to discriminate between out-tol and in-tol points to obtain a the mean value and standard deviation of the function at a point that is not in the training data-set. This forward inferencing of the FF NN for regression is illustrated in figure \ref{fig:predictionphase}, wherein the value of the function is obtained at $x_{\text{query}}\left(\notin \text{training data-set}\right)$. First, several trial points are generated along the $Y$-axis with the $X$ value as $x_{\text{query}}$. Then we classify these trial points as either in-tol or out-tol using a forward pass with both labels-- 1 (in-tol) and 0 (out-tol) for all the trial points. Trial points which possess higher goodness for label 1 can be considered in-tol, and those with higher goodness for label 0 can be considered out-tol (see discussuion in \ref{subsubsec:goodnessinvert}). Next, the mean value for $y$ and the $95\%$ confidence interval ($\pm$ twice the standard deviation) can be computed using the in-tol points. This process of function value inference is viable for functions of multiple variables as well, with $x_{\text{query}}$ being multidimensional. This process of obtaining $y_{\text{mean}}$ can be repeated over multiple $X_{\text{query}}$ points to obtain a smoother curve for the function. Algorithm \autoref{alg:forward_forward_prediction} details function regression using a FF NN at a point $x_{\text{query}}$.

\begin{algorithm}[H]
\small
    \caption{Forward Forward Regression Training for 1-D functions}
    \begin{algorithmic}[1]
        \REQUIRE \ \\ 
        Training dataset: $\mathcal{D}_{\text{train}}=\left\{\left(x_{\text{actual}}^{(i)},y_{\text{actual}}^{(i)}\right):i\in\{1,2,\dots,N_{\text{actual}}\}\right\}$ \\
        Preset tolerance level: tol \\

        \item[] 
        \STATE Define a Forward Forward NN that takes 3 inputs-- $x$ coordinate, $y$ coordinate and label (either 1.0 or 0.0). The NN has $N_{\text{layers}}$ number of layers, with parameters $\theta_{(i)}$ associated with the $i^{\text{th}}$ layer, and the output of the layer being $\boldsymbol{y}^{(i)}_{\theta_{(i)}}$.
        \STATE Define arbitrary vectors $\boldsymbol{\zeta}^{(i)}$ where $ i\in\{1,2,\dots,N_{\text{layers}}\}$.
        \STATE Define layer-wise goodness functions $g^{(i)}\left(\boldsymbol{y}^{(i)}_{\theta_{(i)}}\right) = cos_{sim}\left(\boldsymbol{y}^{(i)}_{\theta_{(i)}},\zeta^{(i)}\right)$
        \STATE Define $y_{\text{min}}$ and $y_{\text{max}}$ as feasible limits for the range of the function $y(x)$ in the domain of interest.

        \FOR{$i=1,\dots,N_{\text{actual}}$}

                \STATE $\mathcal{D}_{\text{train}} = \mathcal{D}_{\text{train}} \cup \{(x_{\text{actual}}^{(i)},y_{i}^{(k)}):k\in{1,\dots,N_{\text{in-tol}}}$\}
                \\ the points $y_{i}^{(k)}$ are evenly spaced in the interval $\left[y_{\text{actual}}^{(i)}-\text{tol},y_{\text{actual}}^{(i)}+\text{tol}\right]$
                \STATE $\mathcal{D}_{\text{train}} = \mathcal{D}_{\text{train}} \cup \{(x_{\text{actual}}^{(i)},y_{i}^{(k)}):k\in{1,\dots,N_{\text{out-tol}}}$\}
                \\ the points $y_{i}^{(k)}$ are evenly spaced in the interval $\left[y_{\text{min}},y_{\text{actual}}^{(i)}-\text{tol}\right)\cup\left(y_{\text{actual}}^{(i)}+\text{tol},y_{\text{max}}\right]$

        \ENDFOR

         \STATE Define correctly labelled (positive) dataset as follows: \\ $\mathcal{D}_{\text{positive}} = \left\{(x_{\text{actual}}^{(i)}, y_i,1.0):\left|y_{\text{actual}}^{(i)}-y_i\right|\leq \text{tol}\right\} \cup \left\{(x_i, y_i,0.0):\left|y_{\text{actual}}^{(i)}-\text{tol}\right|>\text{tol}\right\}$
         \item[] Define incorrectly labelled (negative) dataset as follows: \\ $\mathcal{D}_{\text{negative}} = \left\{(x_{\text{actual}}^{(i)}, y_i,0.0):\left|y_{\text{actual}}^{(i)}-y_i\right|\leq \text{tol}\right\} \cup \left\{(x_{\text{actual}}^{(i)}, y_i,1.0):\left|y_{\text{actual}}^{(i)}-y_i\right|>\text{tol}\right\}$\\
         \item[] Where $(x_{\text{actual}}^{(i)},y_i)\in\mathcal{D}_{\text{train}} \, \forall \, i\in{1,\dots,N_{\text{actual}}}$

         \STATE Define inputs $\boldsymbol{\xi}_{\text{pos}}^{(0)} = \mathcal{D}_{\text{positive}}$ and $\boldsymbol{\xi}_{\text{neg}}^{(0)} = \mathcal{D}_{\text{negative}}$
        \FOR{i=$1,\dots,N_{\text{layers}}$}
        \FOR{epoch = $1,\dots,N_{\text{epochs}}$}
        \STATE Perform forward pass through layer $i$ with inputs as $\boldsymbol{\xi}_{\text{pos}}^{(i-1)}$, $\boldsymbol{\xi}_{\text{neg}}^{(i-1)}$ to obtain outputs $\boldsymbol{y}^{(i)}_{\text{pos},\theta_{(i)}}$, $\boldsymbol{y}^{(i)}_{\text{neg},\theta_{(i)}}$, respectively.
        \STATE  Obtain $g^{(i)}_{\text{pos}}$ and $g^{(i)}_{\text{neg}}$ from $\boldsymbol{y}^{(i)}_{\text{pos},\theta_{(i)}}$ and $\boldsymbol{y}^{(i)}_{\text{neg},\theta_{(i)}}$, respectively, as explained in step 3.
        \STATE Compute mean $Loss^{(i)}$ across all datapoints (Refer Eq. \ref{eqn:Loss_i})
        \STATE Update $\theta_{(i)}$ to minimize $Loss^{(i)}$ (Any gradient descent will do)
        \ENDFOR
        \STATE Set $\boldsymbol{\xi}_{\text{pos}}^{(i)}=\boldsymbol{y}^{(i)}_{\text{pos},\theta_{(i)}}$ and $\boldsymbol{\xi}_{\text{neg}}^{(i)}=\boldsymbol{y}^{(i)}_{\text{neg},\theta_{(i)}}$
        \ENDFOR

        \RETURN Final trained NN model with parameters $\theta_{(i)}$ and arbitrary vectors $\boldsymbol{\zeta}^{(i)}$ for $i=1,\dots,N_{\text{layers}}$
    \end{algorithmic}
    \label{alg:training}
\end{algorithm}

\begin{algorithm}[H]
\small
    \caption{Forward-Forward prediction of function value at $x_{\text{query}}$}
    \begin{algorithmic}[1]
        \REQUIRE \ \\Trained FF NN model with $N_{\text{layers}}$ number of layers and similar number of arbitrary vectors $\boldsymbol{\zeta}^{(i)}$. \\
        The $x$-coordinate at which function ($y$) value is desired: $x_{\text{query}}$. \\
        An estimate of the upper ($y_{\text{max}}$) and lower ($y_{\text{min}}$) limit of the function's range
        Number of trial points to be generated at each query point: $N_{\text{trials}}$
     
        \STATE Initialize $G_{\text{in-tol}}\gets$ zeros($N_{\text{trials}}$,1)
        \STATE Initialize $G_{\text{out-tol}}\gets$ zeros($N_{\text{trials}}$,1)
        \STATE Initialize $\xi^{(0)}_{\text{in-tol}} \gets$ zeros($N_{\text{trials}}$,3)
        \STATE Initialize $\xi^{(0)}_{\text{out-tol}} \gets$ zeros($N_{\text{trials}}$,3)
        \FOR{$k=1,\dots,N_{\text{trials}}$}
            \STATE Define $y^{(k)}_{\text{trial}}=y_{\text{min}}+\frac{y_{\text{max}}-y_{\text{min}}}{N_{\text{trials}}-1}(k-1)$
            
            \STATE $\xi^{(0)}_{\text{in-tol}}[k] \gets (x_{\text{query}},y^{(k)}_{\text{trial}},1.0)$
            \STATE $\xi^{(0)}_{\text{out-tol}}[k] \gets (x_{\text{query}},y^{(k)}_{\text{trial}},0.0)$
        \ENDFOR
        \FOR{$i=1,\dots,N_{\text{layers}}$}
            \STATE Input $\xi^{(i-1)}_{\text{in-tol}}$ to the $i^{th}$ layer of the FF NN to obtain $y^{(i)}_{\theta_i,\text{in-tol}}$ as the output.
            \STATE Input $\xi^{(i-1)}_{\text{out-tol}}$ to the $i^{th}$ layer of the FF NN to obtain $y^{(i)}_{\theta_i,\text{out-tol}}$ as the output.
            \STATE Compute $g^{(i)}_{\text{in-tol}} \gets cos_{sim}\left(y^{(i)}_{\theta_i,\text{in-tol}},\boldsymbol{\zeta}^{(i)}\right)$
            \STATE Compute $g^{(i)}_{\text{out-tol}} \gets cos_{sim}\left(y^{(i)}_{\theta_i,\text{out-tol}},\boldsymbol{\zeta}^{(i)}\right)$
            \STATE $G_{\text{in-tol}} \gets G_{\text{in-tol}}+ g^{(i)}_{\text{in-tol}}$
            \STATE $G_{\text{out-tol}} \gets G_{\text{out-tol}}+ g^{(i)}_{\text{out-tol}}$
            \STATE $\xi^{(i)}_{\text{in-tol}} \gets y^{(i)}_{\theta_i,\text{in-tol}}$
            \STATE $\xi^{(i)}_{\text{out-tol}} \gets y^{(i)}_{\theta_i,\text{out-tol}}$
        \ENDFOR
        \STATE Initialize $y \leftarrow \{\}$ (Empty set)
        \FOR{$k=1,\dots,N_{\text{trials}}$}
            \IF{$G_{\text{out-tol}}[k]>G_{\text{in-tol}}[k]$}
            \STATE $y \gets y \cup \{y^{(k)}_{\text{trial}}\}$
            \ENDIF
        \ENDFOR
        \STATE Define $y_{\text{mean}} \gets mean(y)$
        \STATE Define $y_{\text{std}} \gets STD(y)$
        \RETURN $y_{\text{mean}}$ and $y_{\text{std}}$ to obtain 95\% confidence interval ($\pm2y_{\text{std}}$) 
   
    \end{algorithmic}
    \label{alg:forward_forward_prediction}
\end{algorithm}

\section{Results and discussions}
\label{sec:Results}
We validated our proposed FF regression algorithm against various benchmark 1-D, 2-D and 3-D functions. We chose functions involving combinations of the ubiquitous sinusoidal and exponential terms. Across all the regression tasks considered in the study we used a similar FF NN (dimension of input varies) with a total of 3 layers with 64, 128 and 32 neurons in each layer, respectively. We employ the GELU activation function in each layer. Further details regarding the hyperparameters employed in each benchmark is available in Table \ref{tab:Hyperparameters}.
\subsection{1-D regression}
\begin{figure}[htbp]
    \centering
    \begin{subfigure}[b]{0.23\textwidth}
        \includegraphics[width=\textwidth]{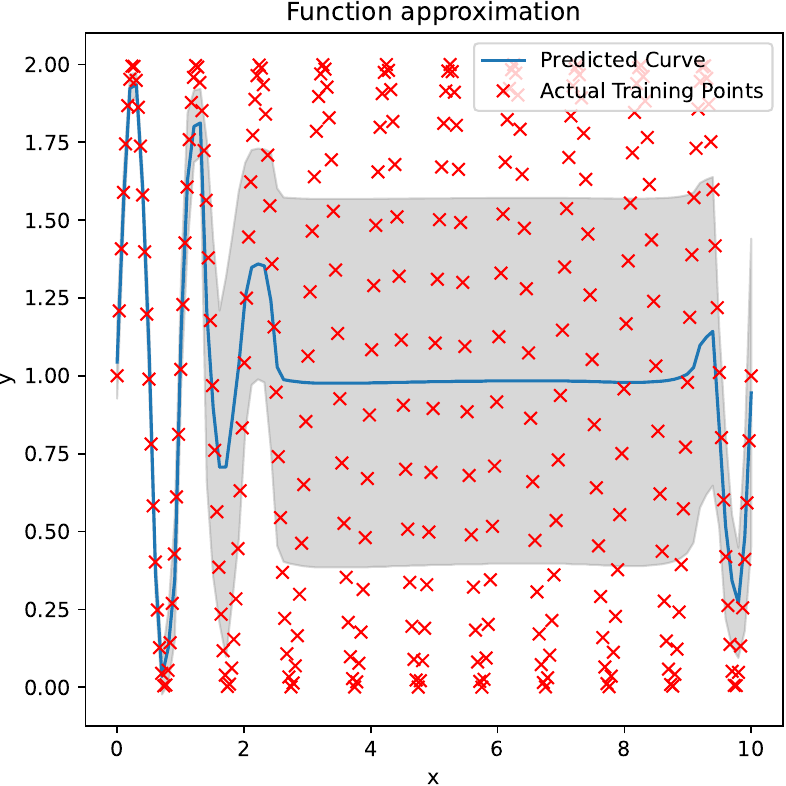}
        \caption{}
        \label{fig:periodic}
    \end{subfigure}
    \begin{subfigure}[b]{0.23\textwidth}
        \includegraphics[width=\textwidth]{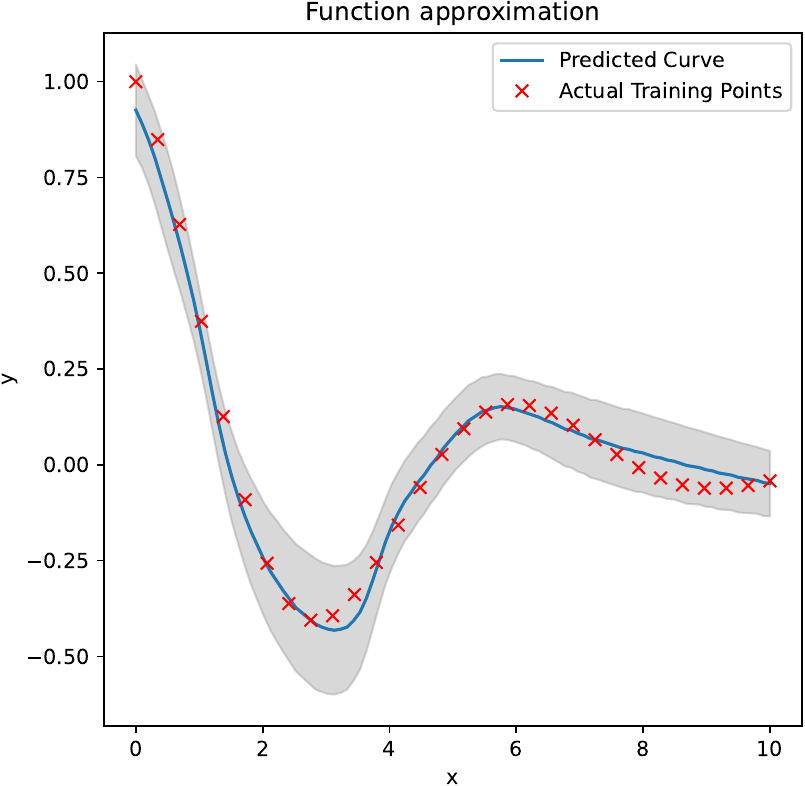}
        \caption{}
        \label{fig:damped}
    \end{subfigure}
    \begin{subfigure}[b]{0.23\textwidth}
        \includegraphics[width=\textwidth]{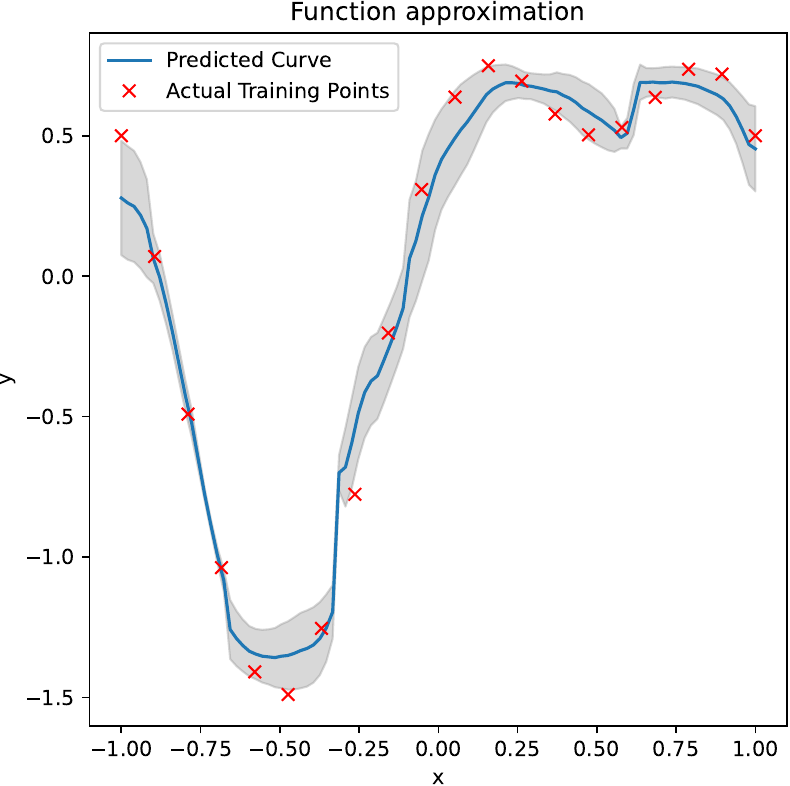}
        \caption{}
        \label{fig:base}
    \end{subfigure}
    \caption{Results of FF-regrssion on 1D functions-- (a) $f_1(x)$, (b) $f_2(x)$ and (c) $f_3(x)$, with the red crosses indicating the training data points, blue line indicating the mean predicted value, while the shaded area denotes the 95\% confidence interval.}
    \label{fig:three_functions}
\end{figure}

In figure \ref{fig:three_functions} we provide a summary of FF-regression results for three different functions:
\begin{itemize}
    \item $f_1(x) = sin(2\pi x) + 1$
    \item $f_2(x) = e^{-0.3 x}cos(\frac{\pi x}{2})$
    \item $f_3(x) = sin(\pi x)+\frac{1}{2}cos(2\pi x)$
\end{itemize}

The plots provided show the training data points as red crosses, and the mean predicted value ($y_{\text{mean}}$) of the function as a blue curve with the shaded area denoting the 95\% confidence region. Each of the plots show the predictions at their ``best" with convergence evaluated after increasing the number of training data points and increasing the number of epochs of training per layer. As expected, increasing the number of training points and the number of training epochs show improvement in accuracy and reduction in uncertainty to an extent.
In figure \ref{fig:periodic} we notice that FF NN is unable to approximate all cycles of the sinusoid despite a large number of training datapoints (300) and training epochs (5000), presumably for want of more complexity in the NN. However, the other FF NNs are able to approximate the other functions (figures \ref{fig:damped},\ref{fig:base}) ,which contain 1-2 full period cycles of the function, very accurately with around 20 data points and 500 epochs of training.
\subsubsection{Comment on varying the hyperparameters}
The effect of the following hyperparameters from Algorithms \ref{alg:training} and \ref{alg:forward_forward_prediction} were studied:
\begin{itemize}
    \item tol: It was observed that decreasing the tol parameter improved the accuracy and reduced the uncertainty in the predicted results. However, too small a ``tol" can result in breaks in the function prediction, wherein the entire set of trial points would be classified as out-tol. An example of this can be seen in figure \ref{fig:tolstudy}
    \item $y_{\text{min}}$ and $y_{\text{max}}$: Figure \ref{fig:ymin} shows that as $y_{\text{min}}$ gets too close to the least $y_{\text{actual}}$, the FF NN provides poor prediction at such points as enough number of trial points are not generated in the interval $[y_{\text{min}},y_{\text{actual}}-tol]$.
    \item $N_{\text{in-tol}}$ and $N_{\text{out-tol}}$: During training, at any given $x_{\text{actual}}$ the length of the in-tol region is clearly smaller than the length of the out-tol region. This would mean that we would need more number of out-tol  points as compared to in-tol points, i.e, $N_{\text{out-tol}}$ should be significantly greater than $N_{\text{in-tol}}$. Figure \ref{fig:intolouttol} provides a plot of MSE vs. $N_{\text{out-tol}}$ for FF-regression of $f_3$, showing that higher $N_{\text{out-tol}}$ as compared to $N_{\text{in-tol}}$ provides more accuracy.
\end{itemize}
The hyperparameters relevant to the FF-regression for each of the functions considered in the study is summarized in Table \ref{tab:Hyperparameters}.
\subsubsection{Comment on inversion of Goodness}
\label{subsubsec:goodnessinvert}
As discussed in \autoref{subsec:FFregression}, we train the FF NN to increase its goodness value for correctly labeled data and decrease its goodness value for incorrectly labeled data. A peculiar discovery we made during inferencing from a trained FF NN is that the trial points in the vicinity of the in-tol region have a higher goodness score for the out-tol label as opposed to the in-tol label and the trial points away from the in-tol region show higher goodness scores for the in-tol label. This is in stark contrast to the training, where, as shown in figure \ref{fig:layerloss}, $g_{\text{pos}}>g_{\text{neg}}$. This would mean the FF NN is working in a manner exactly opposite to what was intended (which is also useful). This is the reason we use the inequality in line 22 of algorithm \ref{alg:forward_forward_prediction}, wherein we select points with higher goodness for the out-tol label as the in-tol points.
\subsection{2-D and 3-D regression}
\begin{figure}[ht!]
    \centering
    \begin{subfigure}[b]{0.25\linewidth}
        \centering
        \includegraphics[width=\linewidth]{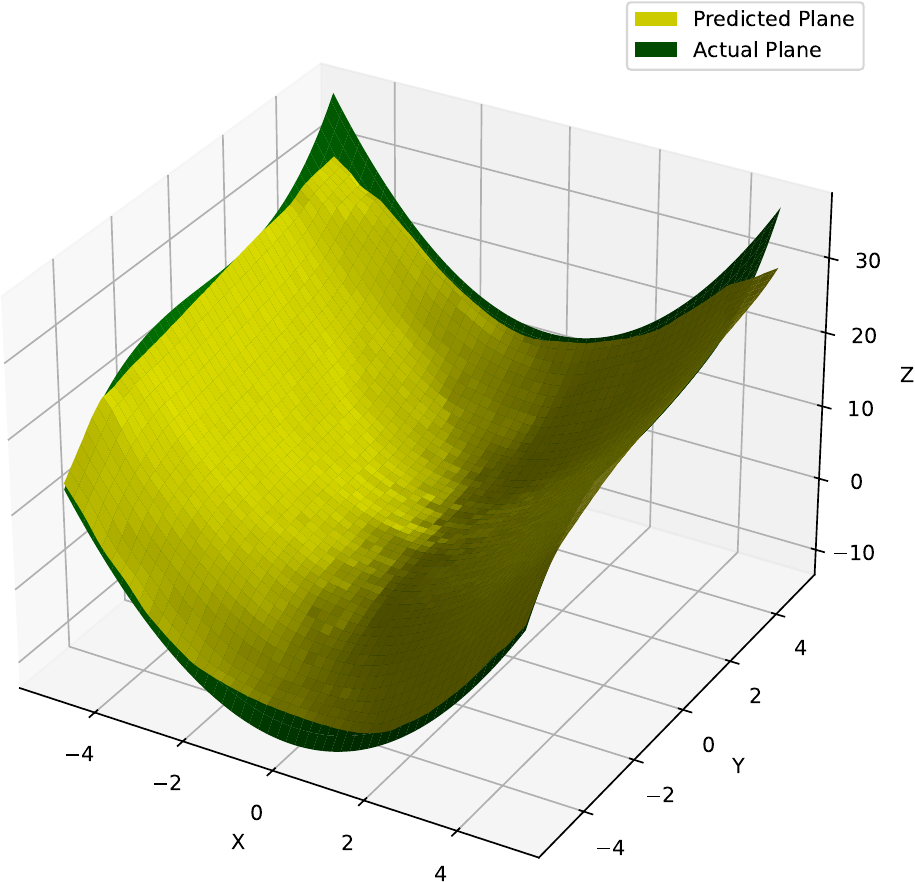}
        \caption{}
        \label{fig:2dregression-a}
    \end{subfigure}
    \begin{subfigure}[b]{0.25\linewidth}
        \centering
        \includegraphics[width=\linewidth]{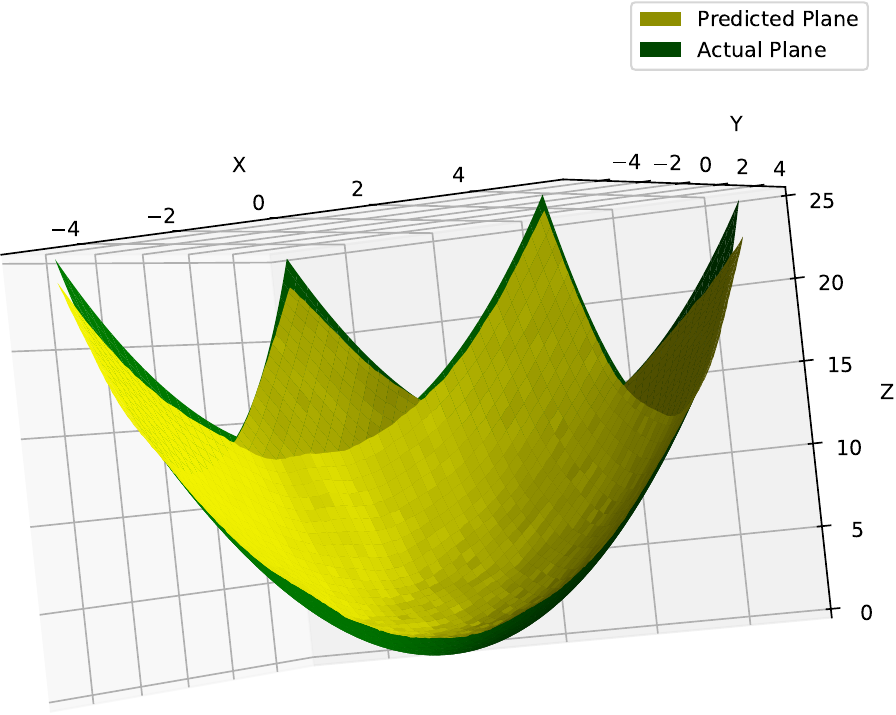}
        \caption{}
        \label{fig:2dregression-b}
    \end{subfigure}
    \caption{FF-Regression results for the 2D functions-- (a) $f_4(x_1,x_2)$ and (b) $f_5(x_1,x_2)$, with the yellow surface indicating the actual function output and the green surface indicating the mean predicted function values. The training datapoints and confidence bounds are omitted for clarity.}
    \label{fig:2dregression}
    \end{figure}

In figure \ref{fig:2dregression} we provide FF-regression results for the 2-D functions:
\begin{itemize}
    \item $f_4(x_1,x_2) = x_1^2+x_2^2$
    \item $f_5(x_1,x_2) = 2 sin(x_1)+cos(x_2)$
\end{itemize}
We omit the uncertainty surfaces for ease of illustration, and the functions can be seen to be approximated reasonably well after 500 epochs of training per layer, with a 25$\times$25 grid of points on the $x_1-x_2$ plane used for training.

\begin{figure}[h]
    \centering
    \begin{subfigure}[c]{0.40\textwidth}
        \centering
        \includegraphics[width=\textwidth]{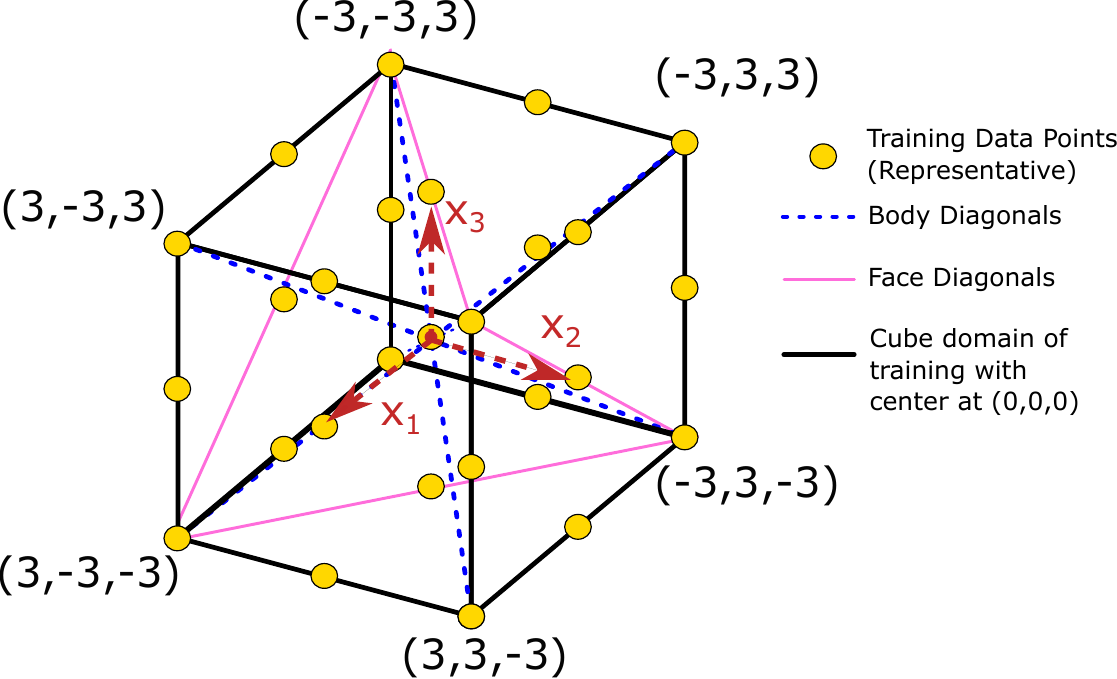}
        \caption{}
        \label{subfig:cube1}
    \end{subfigure}
    \begin{subfigure}[c]{0.48\textwidth}
        \centering
        \includegraphics[width=\textwidth]{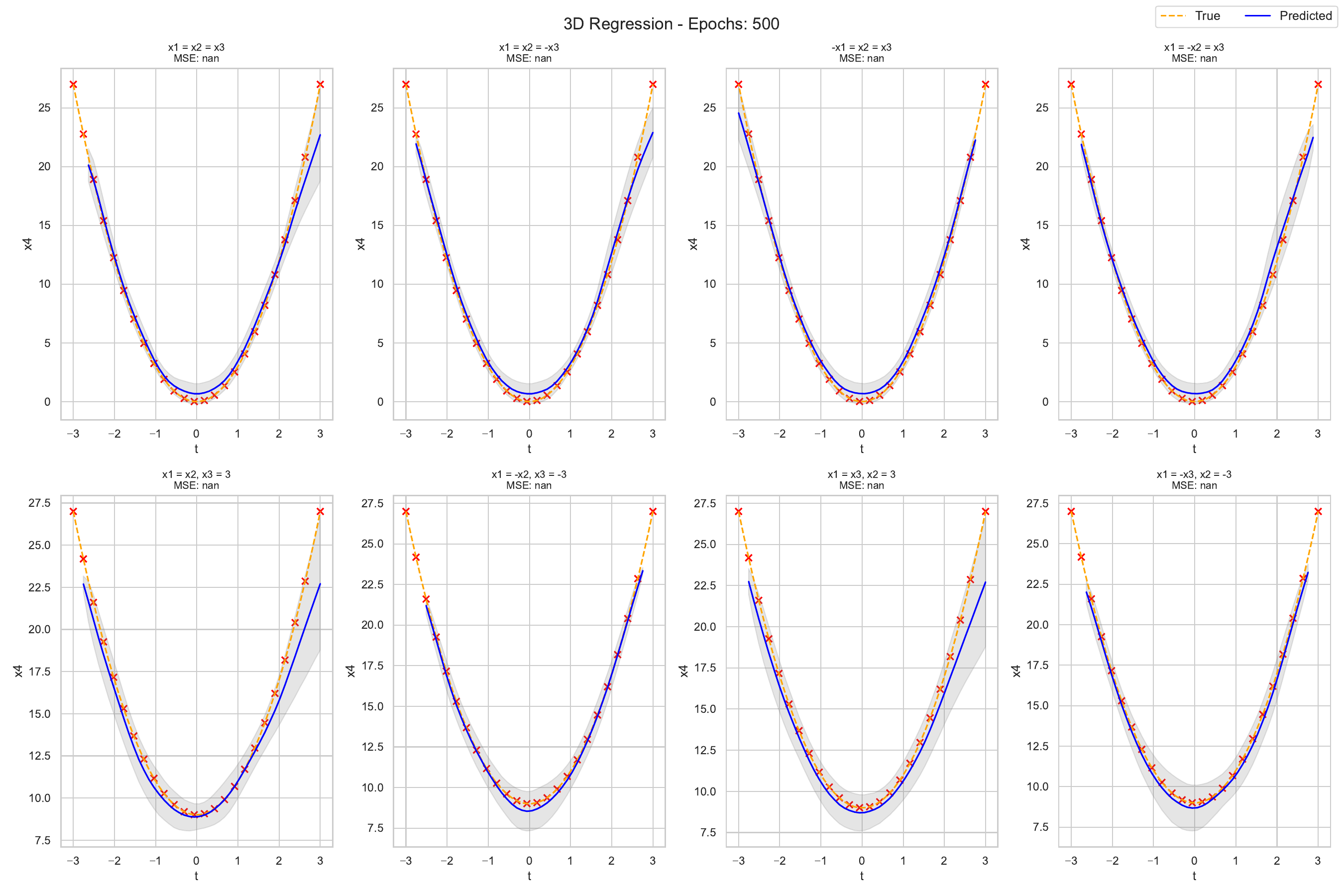}
        \caption{}
    \end{subfigure}
    \caption{(a) Schematic of the domain of training of the FF NN, with the yellow spheres indicating a few of the training data points, the blue-dashed line indicating the 4 body diagonals of the cube and the pink lines indicating the 4 particular surface diagonals along which the true and predicted line plots were compared. (b) Line plots of the FF-Regression result for 3D function $f_6$, with the red crosses indicating training datapoints and the gray shading indicating the 95\% confidence region.}
    \label{fig:fun6}
\end{figure}

\begin{figure}[h]
    \centering
    \begin{subfigure}[c]{0.40\textwidth}
        \centering
        \includegraphics[width=\textwidth]{images/cube.pdf}
        \caption{}
        \label{subfig:cube2}
    \end{subfigure}
    \begin{subfigure}[c]{0.48\textwidth}
        \centering
        \includegraphics[width=\textwidth]{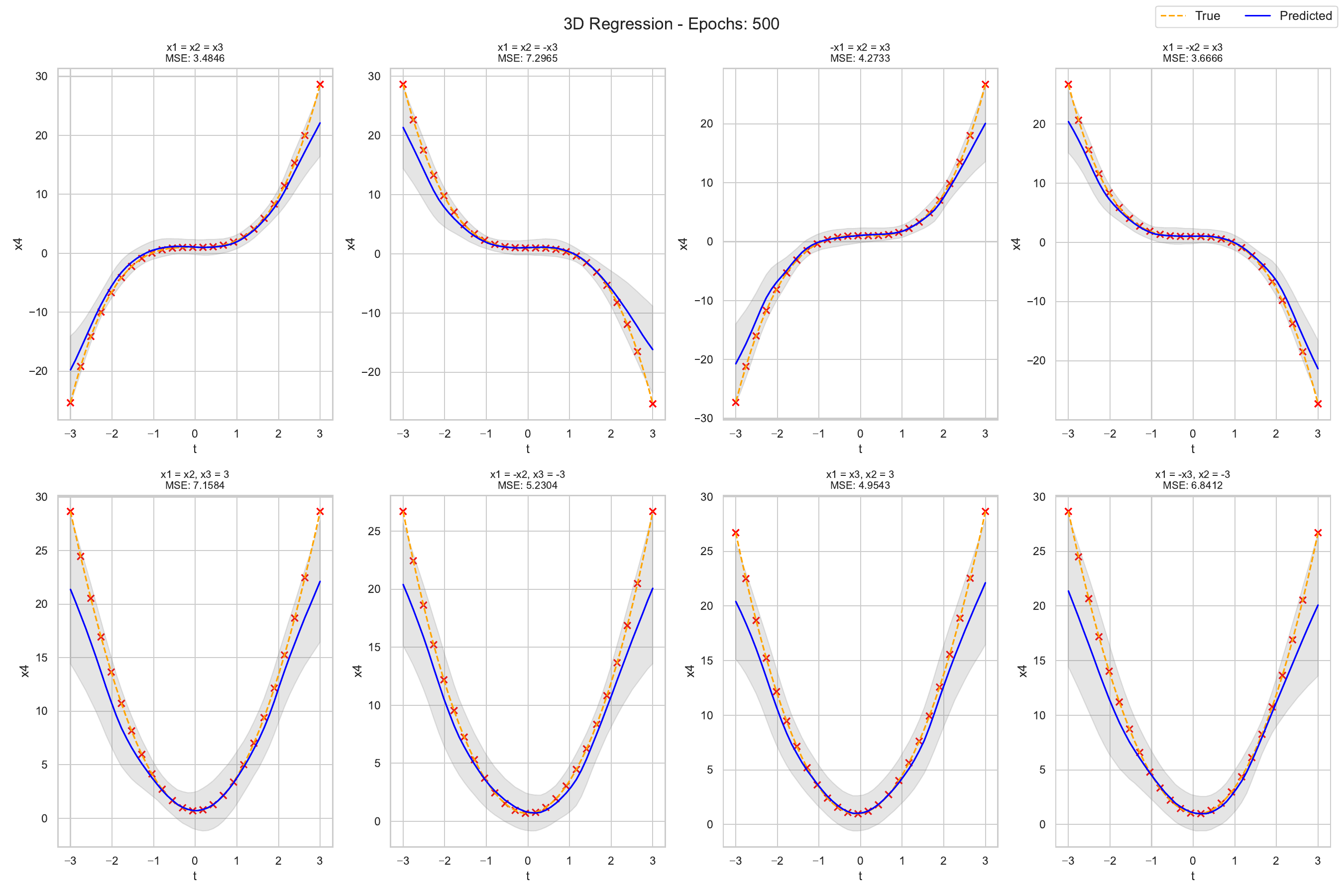}
        \caption{}
    \end{subfigure}
    \caption{(a) Schematic of the domain of training of the FF NN (same as for figure \ref{subfig:cube1}). (b) Line plots of the FF-Regression result for 3D function $f_7$, with the red crosses indicating training datapoints encountered along the line plot and the gray shading indicating the 95\% confidence region.}
    \label{fig:fun7}
\end{figure}

\begin{figure}[b!]
    \centering
    \begin{subfigure}[c]{0.40\textwidth}
        \centering
        \includegraphics[width=\textwidth]{images/cube.pdf}
        \caption{}
        \label{subfig:cube3}
    \end{subfigure}
    \begin{subfigure}[c]{0.48\textwidth}
        \centering
        \includegraphics[width=\textwidth]{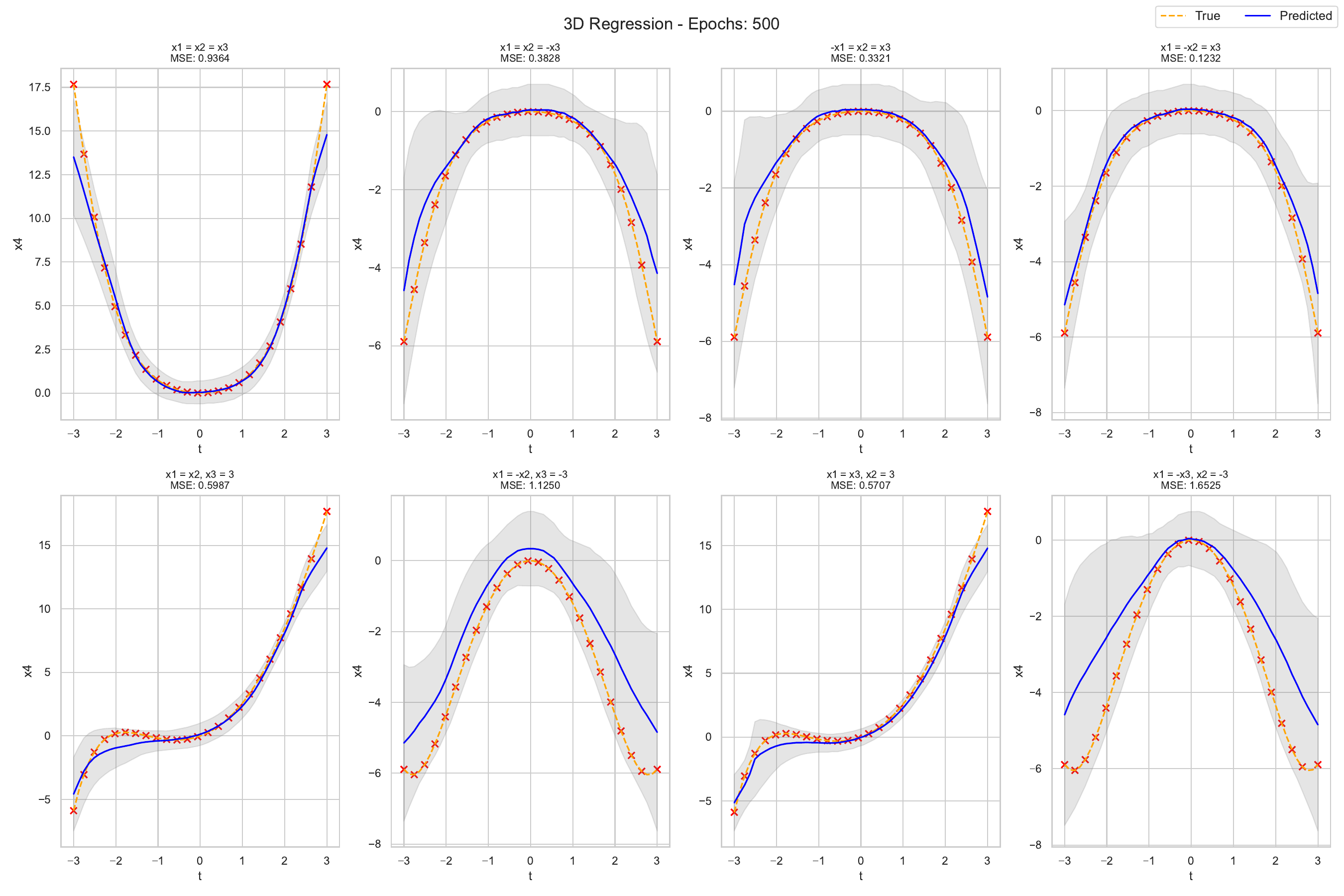}
        \caption{}
    \end{subfigure}
    \caption{(a) Schematic of the domain of training of the FF NN ((same as for figure \ref{subfig:cube1})). (b) Line plots of the FF-Regression result for 3D function $f_8$, with the red crosses indicating training datapoints encountered along the line plot and the gray shading indicating the 95\% confidence region.}
    \label{fig:fun8}
\end{figure}

We chose the following 3-D functions as the next benchmark for our proposed FF-regression algorithm:
\begin{itemize}
    \item $f_6 (x_1,x_2,x_3) = x_1^2+x_2^2+x_3^2$
    \item $f_7 (x_1,x_2,x_3) = sin(\frac{x_1 x_2}{5})+cos^2(\frac{x_3}{5})+x_1x_2x_3$
    \item $f_8 (x_1,x_2,x_3) = e^{\frac{x_1^2}{5}}sin(\frac{x_2x_3}{5})+e^{\frac{x_2^2}{5}}sin(\frac{x_1x_3}{5})+e^{\frac{x_3^2}{5}}sin(\frac{x_2x_1}{5})$
\end{itemize}
We used 25$\times$25$\times$25 grid of training points distributed within a cubical domain with $x_1,x_2,x_3 \in [-3,3]$. Visualizing this result would require a 4-D plot. Instead, noting that the domain of the above 3-D functions is contained within a cube, we chose to compare the true and predicted functions along certain lines in the domain. As illustrated in figures \ref{subfig:cube1}, \ref{subfig:cube2} and \ref{subfig:cube3}, we chose the 4 body diagonals and 4 other surface diagonals on the cube to compare the true and predicted values of the functions. The FF-regression results are shown in figures \ref{fig:fun6},\ref{fig:fun7} and \ref{fig:fun8}, with 500 epochs of training providing satisfactory accuracy for all functions. Increasing the number of epochs further seems to marginally improve the accuracy and the uncertainty. It can be noted that the output corresponding to $f_8$ shows an unusually high uncertainty along certain lines of data. The uncertainty in $f_8$ can be expected to reduce with an increase in the number of training points.


\section{Conclusion}
\label{sec:conc}
In this work we proposed and validated a new algorithm for function regression using the Forward Forward method of training NNs. We successfully benchmarked the proposed algorithm against eight 1-D,2-D and 3-D functions. We documented the effect of various hyperparameters on the accuracy and uncertainty of the predictions. In \autoref{subsubsec:goodnessinvert} we also noted a peculiarity wherein the trained FF NN works exactly opposite to its initial design, thereby still being able to perform function regression. We did not explore the underlying mathematical reason in this work. In \autoref{app:KANs} and \autoref{app:pnn} we provide preliminary results on extending the FF-regression algorithm to Kolmogorov Arnold Networks and Deep Physical Neural Networks, respectively. As seen in \autoref{tab:ffbpcompare}, the traditional backpropagation algorithm significantly outperforms the proposed FF-regression algorithm in terms of compute time to achieve similar accuracies. However, further studies have to be performed to ascertain if the FF-regression algorithm consumes significantly lower energy compared to BP when deployed on a fully Analog/Physical Neural Network framework.

\section{REPRODUCIBILITY STATEMENT}
We have taken all measures to ensure that all the figures and results provided in the main text and appendix are reproducible. The algorithm underlying the training and inferencing of a Forward Forward Neural Network for regression are provided in Algorithms \ref{alg:training} and \ref{alg:forward_forward_prediction} respectively. Furthermore, all codes and information related to the results in the manuscript are available upon reasonable request. Due to the use of randomly generated vectors for evaluating cosine similarity with layer outputs, results may slightly vary from the ones presented in the main text.
\bibliography{iclr2026_conference}

\begin{thebibliography}{18}
\providecommand{\natexlab}[1]{#1}
\providecommand{\url}[1]{\texttt{#1}}
\expandafter\ifx\csname urlstyle\endcsname\relax
  \providecommand{\doi}[1]{doi: #1}\else
  \providecommand{\doi}{doi: \begingroup \urlstyle{rm}\Url}\fi

\bibitem[Chen et~al.(2023)Chen, Zhang, Tappertzhofen, Yang, and Valov]{chen2023electrochemical}
Shaochuan Chen, Teng Zhang, Stefan Tappertzhofen, Yuchao Yang, and Ilia Valov.
\newblock Electrochemical-memristor-based artificial neurons and synapses—fundamentals, applications, and challenges.
\newblock \emph{Advanced materials}, 35\penalty0 (37):\penalty0 2301924, 2023.

\bibitem[Griewank(2012)]{griewank2012invented}
Andreas Griewank.
\newblock Who invented the reverse mode of differentiation.
\newblock \emph{Documenta Mathematica, Extra Volume ISMP}, 389400:\penalty0 26, 2012.

\bibitem[Hinton(2022)]{hinton2022forward}
Geoffrey Hinton.
\newblock The forward-forward algorithm: Some preliminary investigations.
\newblock \emph{arXiv preprint arXiv:2212.13345}, 2022.

\bibitem[Kag \& Saligrama(2021)Kag and Saligrama]{kag2021training}
Anil Kag and Venkatesh Saligrama.
\newblock Training recurrent neural networks via forward propagation through time.
\newblock In \emph{International Conference on Machine Learning}, pp.\  5189--5200. PMLR, 2021.

\bibitem[Khosla et~al.(2020)Khosla, Teterwak, Wang, Sarna, Tian, Isola, Maschinot, Liu, and Krishnan]{khosla2020supervised}
Prannay Khosla, Piotr Teterwak, Chen Wang, Aaron Sarna, Yonglong Tian, Phillip Isola, Aaron Maschinot, Ce~Liu, and Dilip Krishnan.
\newblock Supervised contrastive learning.
\newblock \emph{Advances in neural information processing systems}, 33:\penalty0 18661--18673, 2020.

\bibitem[Li et~al.(2023)Li, Song, Wang, Jiang, Yan, Lin, Li, Rao, Barnell, Wu, et~al.]{li2023memristive}
Yunning Li, Wenhao Song, Zhongrui Wang, Hao Jiang, Peng Yan, Peng Lin, Can Li, Mingyi Rao, Mark Barnell, Qing Wu, et~al.
\newblock Memristive field-programmable analog arrays for analog computing.
\newblock \emph{Advanced Materials}, 35\penalty0 (37):\penalty0 2206648, 2023.

\bibitem[Linnainmaa(1970)]{linnainmaa1970representation}
Seppo Linnainmaa.
\newblock \emph{The representation of the cumulative rounding error of an algorithm as a Taylor expansion of the local rounding errors}.
\newblock PhD thesis, Master’s Thesis (in Finnish), Univ. Helsinki, 1970.

\bibitem[Linnainmaa(1976)]{linnainmaa1976taylor}
Seppo Linnainmaa.
\newblock Taylor expansion of the accumulated rounding error.
\newblock \emph{BIT Numerical Mathematics}, 16\penalty0 (2):\penalty0 146--160, 1976.

\bibitem[Lorberbom et~al.(2024)Lorberbom, Gat, Adi, Schwing, and Hazan]{lorberbom2024layer}
Guy Lorberbom, Itai Gat, Yossi Adi, Alexander Schwing, and Tamir Hazan.
\newblock Layer collaboration in the forward-forward algorithm.
\newblock In \emph{Proceedings of the AAAI Conference on Artificial Intelligence}, volume~38, pp.\  14141--14148, 2024.

\bibitem[Mehonic \& Kenyon(2022)Mehonic and Kenyon]{mehonic2022brain}
Adnan Mehonic and Anthony~J Kenyon.
\newblock Brain-inspired computing needs a master plan.
\newblock \emph{Nature}, 604\penalty0 (7905):\penalty0 255--260, 2022.

\bibitem[Momeni et~al.(2023{\natexlab{a}})Momeni, Rahmani, Mall{\'e}jac, Del~Hougne, and Fleury]{momeni2023backpropagation}
Ali Momeni, Babak Rahmani, Matthieu Mall{\'e}jac, Philipp Del~Hougne, and Romain Fleury.
\newblock Backpropagation-free training of deep physical neural networks.
\newblock \emph{Science}, 382\penalty0 (6676):\penalty0 1297--1303, 2023{\natexlab{a}}.

\bibitem[Momeni et~al.(2023{\natexlab{b}})Momeni, Rahmani, Mall{\'e}jac, del Hougne, and Fleury]{momeni2023phyff}
Ali Momeni, Babak Rahmani, Matthieu Mall{\'e}jac, Philipp del Hougne, and Romain Fleury.
\newblock Phyff: Physical forward forward algorithm for in-hardware training and inference.
\newblock In \emph{Machine Learning with New Compute Paradigms}, 2023{\natexlab{b}}.

\bibitem[Scodellaro et~al.(2023)Scodellaro, Kulkarni, Alves, and Schr{\"o}ter]{scodellaro2023training}
Riccardo Scodellaro, Ajinkya Kulkarni, Frauke Alves, and Matthias Schr{\"o}ter.
\newblock Training convolutional neural networks with the forward-forward algorithm.
\newblock \emph{arXiv preprint arXiv:2312.14924}, 2023.

\bibitem[Sharma et~al.(2024)Sharma, Rath, Kundu, Korkmaz, S, Thompson, Bhat, Goswami, Williams, and Goswami]{sharma2024linear}
Deepak Sharma, Santi~Prasad Rath, Bidyabhusan Kundu, Anil Korkmaz, Harivignesh S, Damien Thompson, Navakanta Bhat, Sreebrata Goswami, R~Stanley Williams, and Sreetosh Goswami.
\newblock Linear symmetric self-selecting 14-bit kinetic molecular memristors.
\newblock \emph{Nature}, 633\penalty0 (8030):\penalty0 560--566, 2024.

\bibitem[Wright et~al.(2022)Wright, Onodera, Stein, Wang, Schachter, Hu, and McMahon]{wright2022deep}
Logan~G Wright, Tatsuhiro Onodera, Martin~M Stein, Tianyu Wang, Darren~T Schachter, Zoey Hu, and Peter~L McMahon.
\newblock Deep physical neural networks trained with backpropagation.
\newblock \emph{Nature}, 601\penalty0 (7894):\penalty0 549--555, 2022.

\bibitem[Wu et~al.(2024)Wu, Xu, Wu, Deng, Xu, Wen, and Li]{wu2024distance}
Yujie Wu, Siyuan Xu, Jibin Wu, Lei Deng, Mingkun Xu, Qinghao Wen, and Guoqi Li.
\newblock Distance-forward learning: Enhancing the forward-forward algorithm towards high-performance on-chip learning.
\newblock \emph{arXiv preprint arXiv:2408.14925}, 2024.

\bibitem[Zhang et~al.(2020)Zhang, Wang, Zhu, Yang, Rao, Song, Zhuo, Zhang, Cui, Shen, et~al.]{zhang2020brain}
Yang Zhang, Zhongrui Wang, Jiadi Zhu, Yuchao Yang, Mingyi Rao, Wenhao Song, Ye~Zhuo, Xumeng Zhang, Menglin Cui, Linlin Shen, et~al.
\newblock Brain-inspired computing with memristors: Challenges in devices, circuits, and systems.
\newblock \emph{Applied Physics Reviews}, 7\penalty0 (1), 2020.

\bibitem[Zolfagharinejad et~al.(2024)Zolfagharinejad, Alegre-Ibarra, Chen, Kinge, and van~der Wiel]{zolfagharinejad2024brain}
Mohamadreza Zolfagharinejad, Unai Alegre-Ibarra, Tao Chen, Sachin Kinge, and Wilfred~G van~der Wiel.
\newblock Brain-inspired computing systems: a systematic literature review.
\newblock \emph{The European Physical Journal B}, 97\penalty0 (6):\penalty0 70, 2024.

\end{thebibliography}
\bibliographystyle{iclr2026_conference}
\newpage
\begin{appendices}
\section{Figures and tables related to the paper}
\begin{figure}[h]
    \centering
    \includegraphics[width=0.4\linewidth]{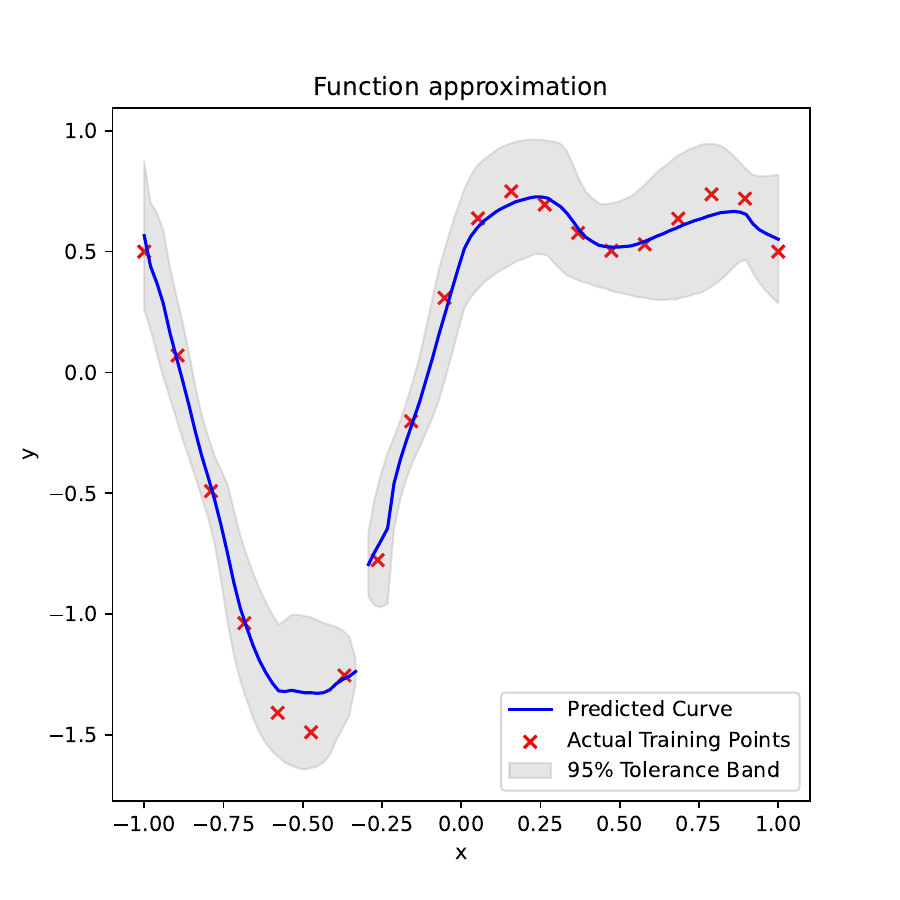}
    \caption{Prediction for function $f_3$ breaks in certain regions as we try to reduce ``tol" below a certain value.}
    \label{fig:tolstudy}
\end{figure}
\begin{figure}[h!]
    \centering
    \begin{minipage}{0.48\linewidth}
        \centering
        \includegraphics[width=\linewidth]{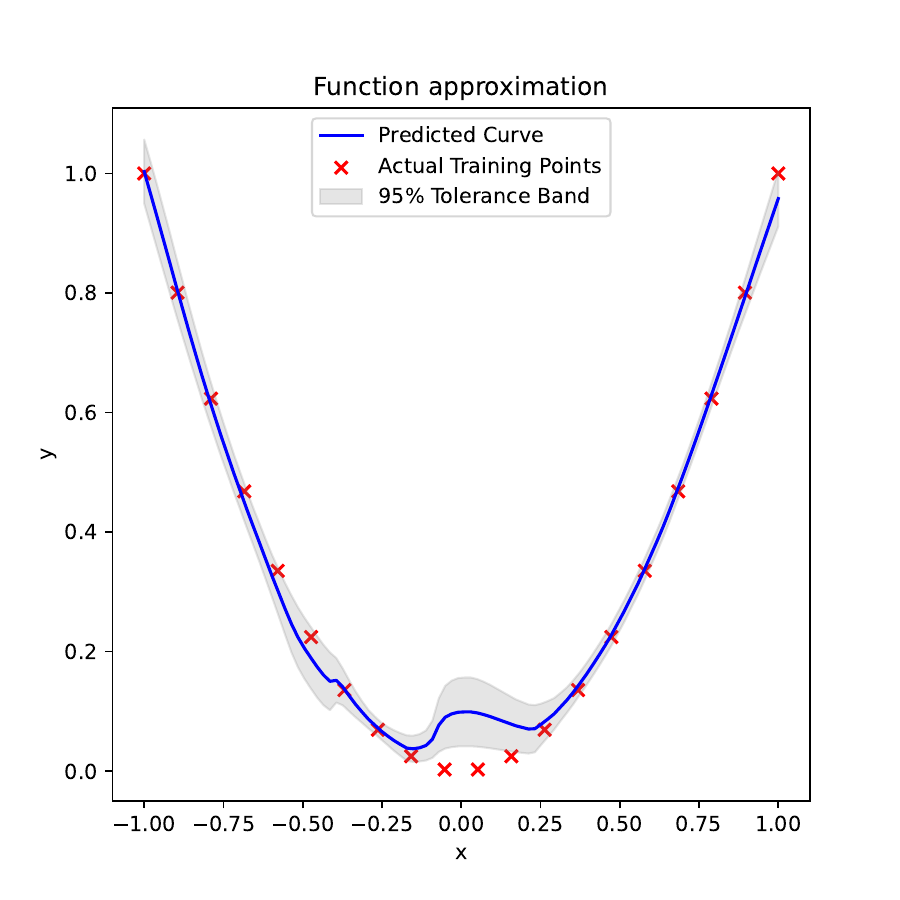}
        \caption*{(a) FF-regression on $y=x^2$ with $y_{\text{min}}=0$.}
        \label{fig:obs1}
    \end{minipage}
    \hfill
    \begin{minipage}{0.48\linewidth}
        \centering
        \includegraphics[width=\linewidth]{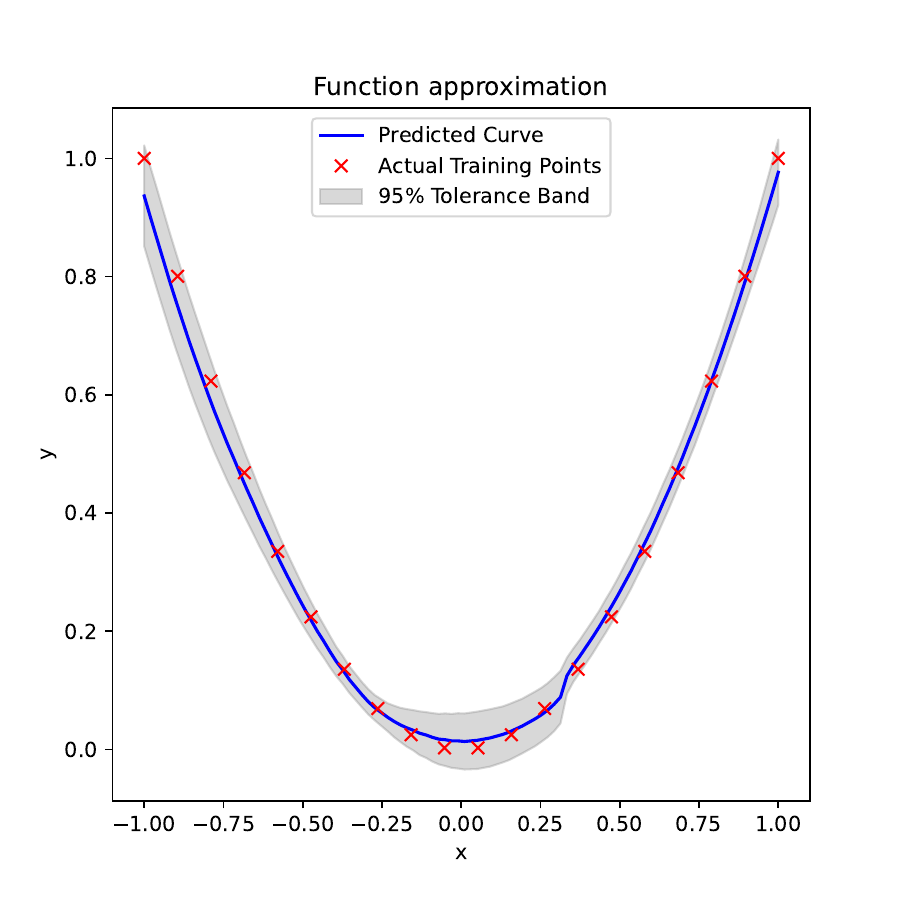}
        \caption*{(b) FF-regression on $y=x^2$ with $y_{\text{min}}=-1$.}
        \label{fig:obs1_true}
    \end{minipage}
    \caption{Comparative study showing that if either $y_{\text{min}}$ or $y_{\text{max}}$ are too close to the training datapoint value ($y_{\text{actual}}$), the FF NN provides poor predictions at such points.}
    \label{fig:ymin}
\end{figure}
\begin{figure}[h!]
    \centering
    \includegraphics[width=0.5\linewidth]{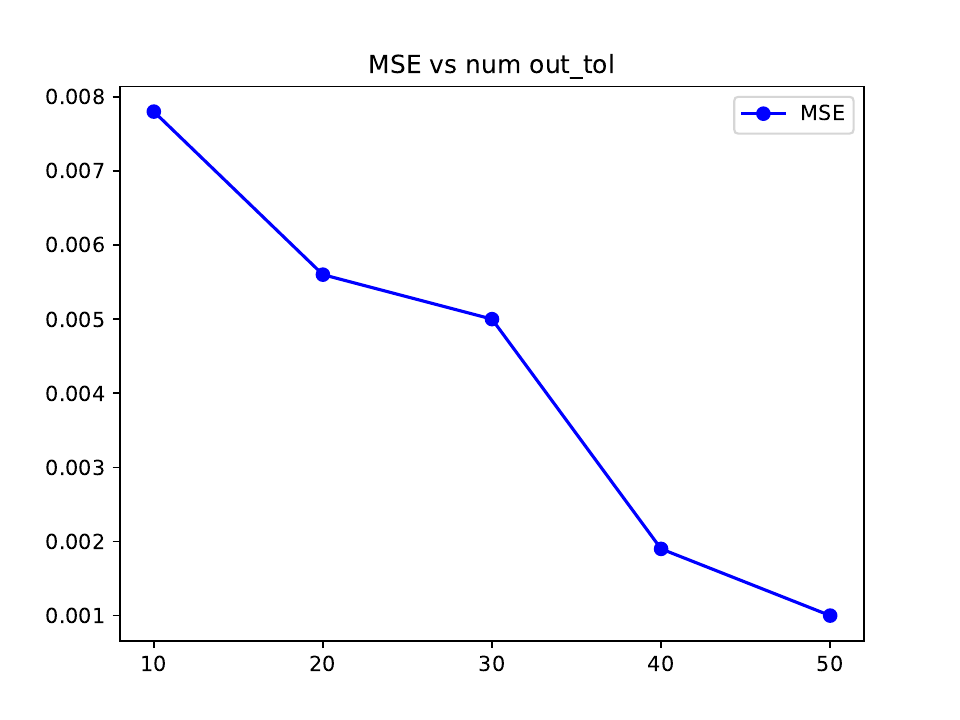}
    \caption{Plot of MSE for FF-regression of $f_3$ as function of $N_{\text{out-tol}}$ used during training.}
    \label{fig:intolouttol}
\end{figure}

\begin{table}[h!]
\centering
\begin{tabular}{|l|c|c|c|c|c|c|c|r|}
\hline

Hyperparameters & $f_1$ & $f_2$ & $f_3$ & $f_4$   & $f_5$  & $f_6$    & $f_7$    & $f_8$ \\
\hline
$N_{\text{layers}}$        & 3         & 3         & 3         & 3               & 3               & 3         & 3         & 3         \\
$\text{tol}$             & 0.02      & 0.05      & 0.01      & 0.1               & 0.1               & 0.1        & 0.1       & 0.1        \\
$N_{\text{in-tol}}$    & 10        & 10        & 10        & 30              & 30              & 30        & 30        & 30        \\
$N_{\text{out-tol}}$     & 10        & 10        & 10        & 50              & 50              & 50        & 50        & 50        \\
$N_{\text{trials}}$    & 1000      & 1000      & 1000      & 300             & 300             & 1000      & 1000      & 1000     \\
$N_{\text{epochs}}$    & 500      & 500      & 500      & 300             & 300             & 500      & 500      & 500     \\
\hline
\end{tabular}
\caption{A list of hyperparameters used for the FF-regression of each function.}
\label{tab:Hyperparameters}
\end{table}

\begin{figure}[h!]
    \centering
    \includegraphics[width=0.6\linewidth]{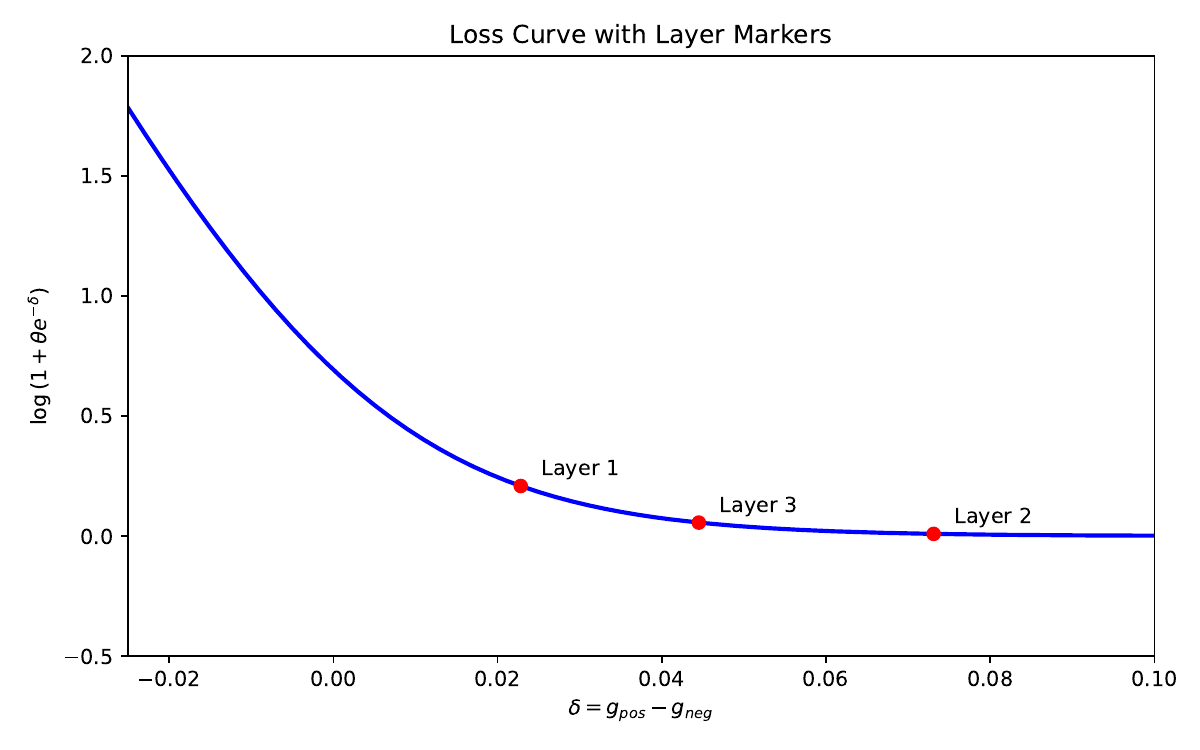}
    \caption{Plot of loss function (Eq. \ref{eqn:Loss_i}) w.r.t $(g_{pos} - g_{neg})$ for layer 1, layer 2 and layer 3 after training for $f_3$.}
    \label{fig:layerloss}
\end{figure}

\begin{table}[h]
\centering
\begin{tabular}{|c|c|c|c|c|}  
\hline
& \begin{tabular}[c]{@{}l@{}}FF Algorithm\\ n\_epochs = 500\end{tabular} & \begin{tabular}[c]{@{}l@{}}Backpropagation\\ n\_epochs = 500\end{tabular} & \begin{tabular}[c]{@{}l@{}}FF Algorithm\\ n\_epochs = 5000\end{tabular} & \begin{tabular}[c]{@{}l@{}}Backpropagation\\ 

n\_epochs = 5000\end{tabular} \\
\hline
$f_3$  & 6.62 s & 0.5 s   & 42.34 s  & 4.67 s      \\
$f_6$   & 173.11 s                         & 0.59 s                     & 2519.19 s                 & 5.01 s      \\
$f_8$   & 155.31 s     & 0.49 s    & 1264.66 s    & 5.10 s      \\
$f_8$ & 305.27 s & 0.52 s   & 2534.62 s   & 5.15 s \\
\hline
\end{tabular}
\caption{Comparison of compute time for NNs with similar number of parameters using BP and FF, for regression of various functions, using a workstation equipped with NVIDIA RTX 5000 Ada Generation, an Intel Xeon w5-2565X CPU (18 cores, 36 threads), and 128 GB of RAM.}
\label{tab:ffbpcompare}
\end{table}

\newpage

\section{Attempts with Kolmogorov Arnold Networks (KANs)}
\label{app:KANs}
\begin{figure}[ht!]
    \centering
    \includegraphics[width=0.5\linewidth]{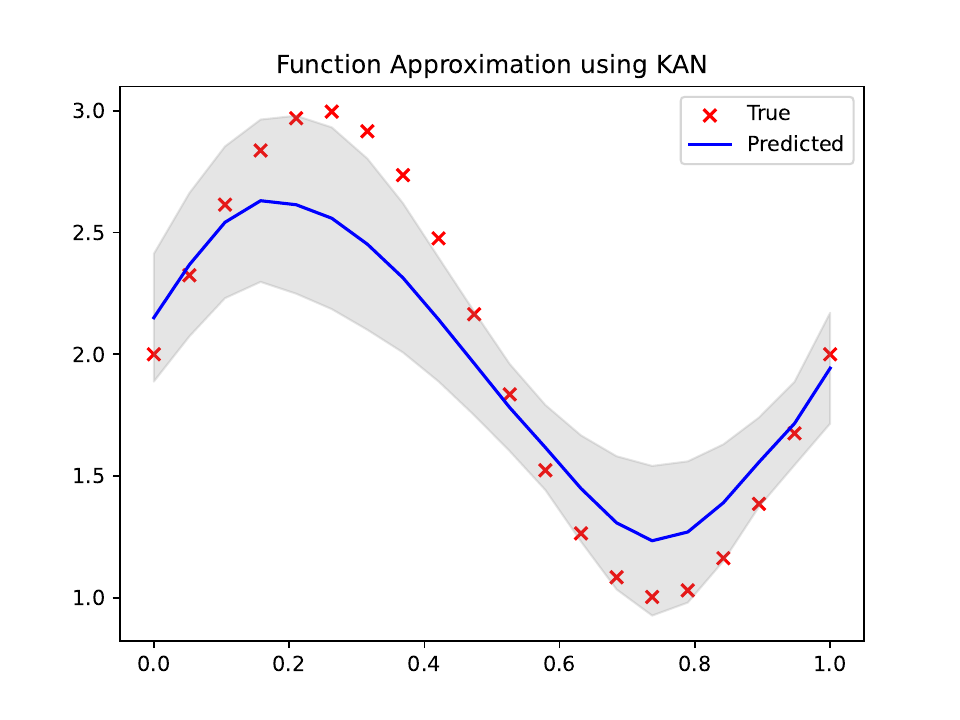}
    \caption{Forward Forward Regression implemented using Kolmogorov Arnold Networks(KANs)}
    \label{fig:kan}
\end{figure}

Kolmogorov Arnold Networks are powerful in approximating complex functions with relatively fewer parameters compared to NNs. Each node of the KAN provides a spline-based approximation and layers of this network act as composite functions. We employed the proposed FF-regression algorithm to train and infer a 3-layered KAN to approximate a simple sinusoidal function $y=2+sin(2\pi x)$. The results after 5000 epochs of training are shown in figure \ref{fig:kan}. This preliminary result seems somewhat encouraging and further studies could provide more insights on effectiveness of training and infering KANs using the FF-regression approach.

\newpage
\section{Results and Discussions for function regression using a Deep Physical Neural Network}
\label{app:pnn}
\begin{figure}[ht!]
    \centering
    % First subfigure
    \begin{subfigure}[b]{0.48\linewidth}
        \centering
        \includegraphics[width=\linewidth]{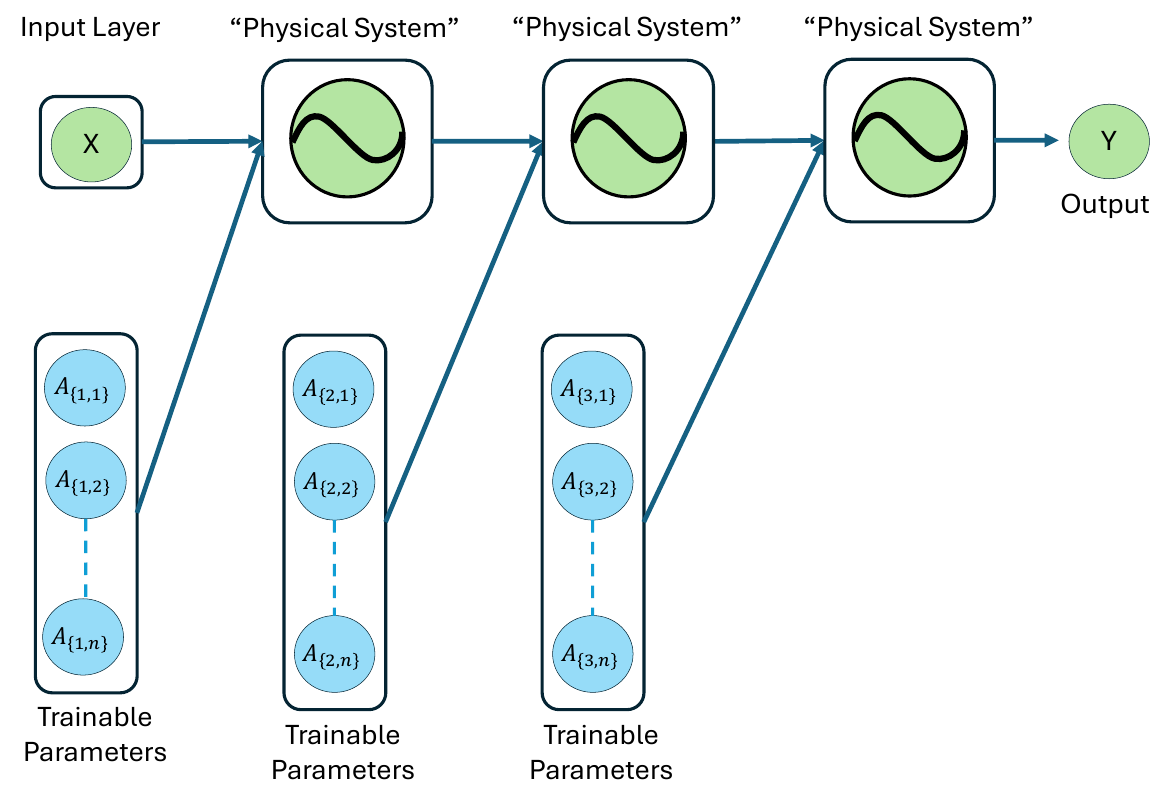}
        \caption{}
        \label{fig:pnnbp}
    \end{subfigure}
    \hfill
    % Second subfigure
    \begin{subfigure}[b]{0.48\linewidth}
        \centering
        \includegraphics[width=\linewidth]{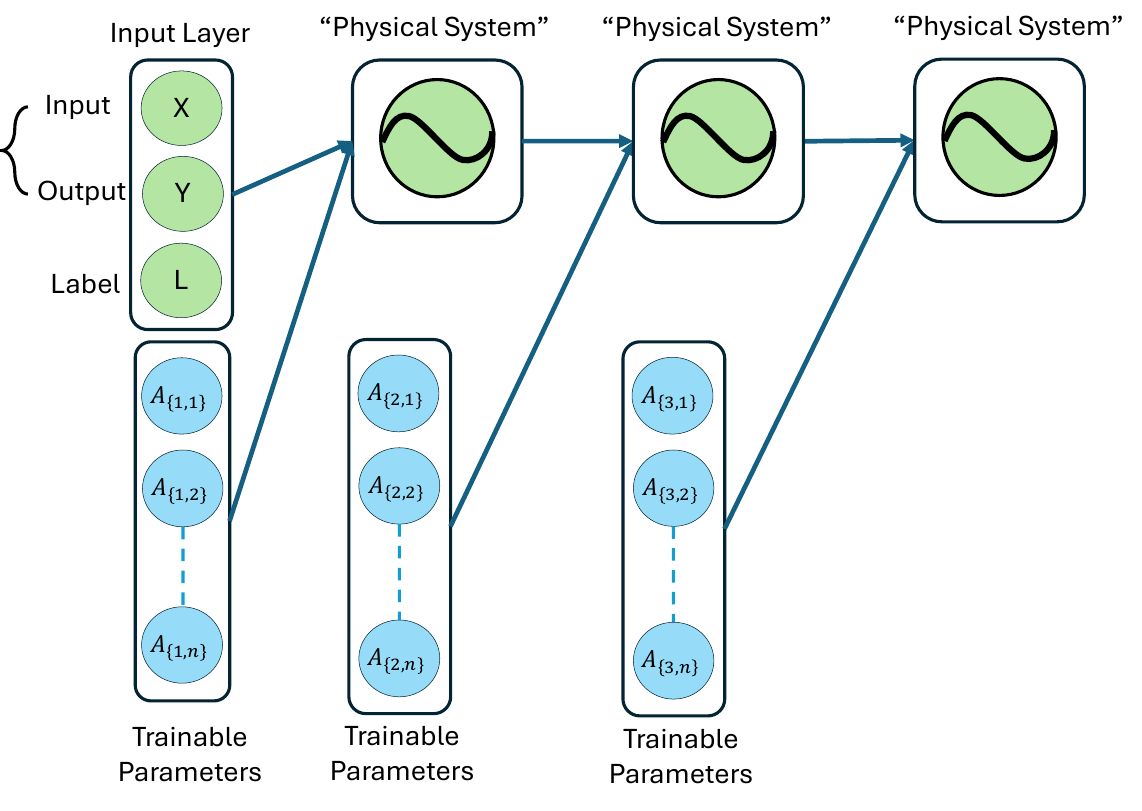}
        \caption{}
        \label{fig:pnnff}
    \end{subfigure}
    
    \caption{Comparison of Deep Physical Neural Networks trained with (a) backpropagation and (b) forward-forward algorithm.}
    \label{fig:pnn_ff_bp}
\end{figure}

We considered a 3 layered DPNN, wherein the trainable parameters are included as part of the input, and the activation function associated with each ``physical layer" is $sin(x)+cos(x)$. We trained a DPNN for regression using the BP algorithm and another DPNN using the FF algorithm. A schematic of the architecture for both can be seen in figure \ref{fig:pnn_ff_bp}.

\begin{figure}[ht!]
    \centering
    \includegraphics[width=0.5\linewidth]{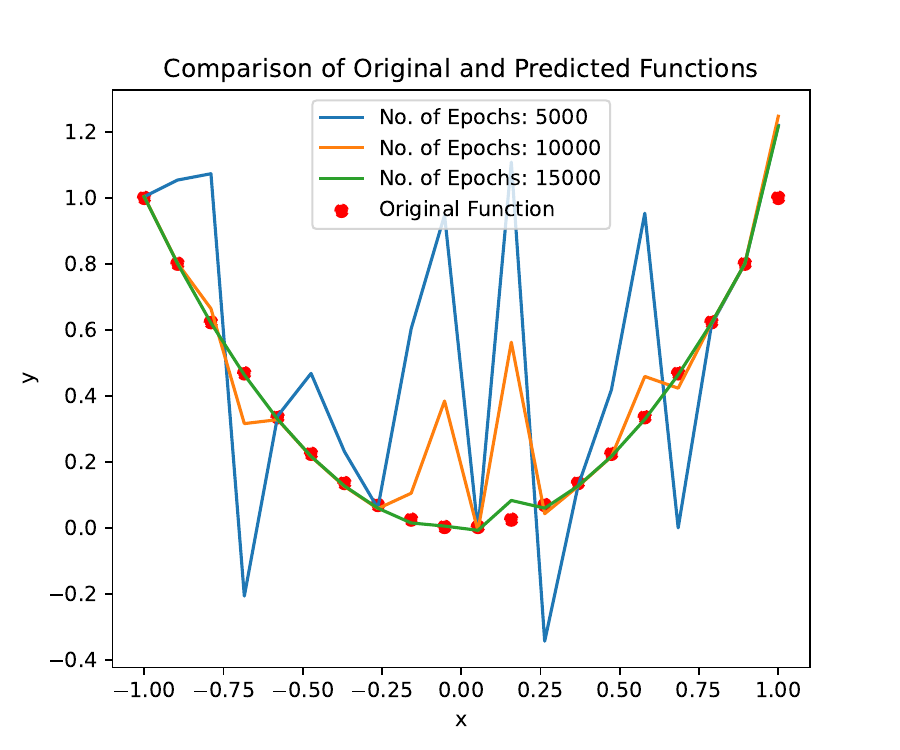}
    \caption{Result after using traditional backpropagation algorithm on physical neural networks with input layer containing trainable parameters.}
    \label{fig:pnnbp_result}
\end{figure}

\begin{figure}[ht!]
    \centering
    \includegraphics[width=0.5\linewidth]{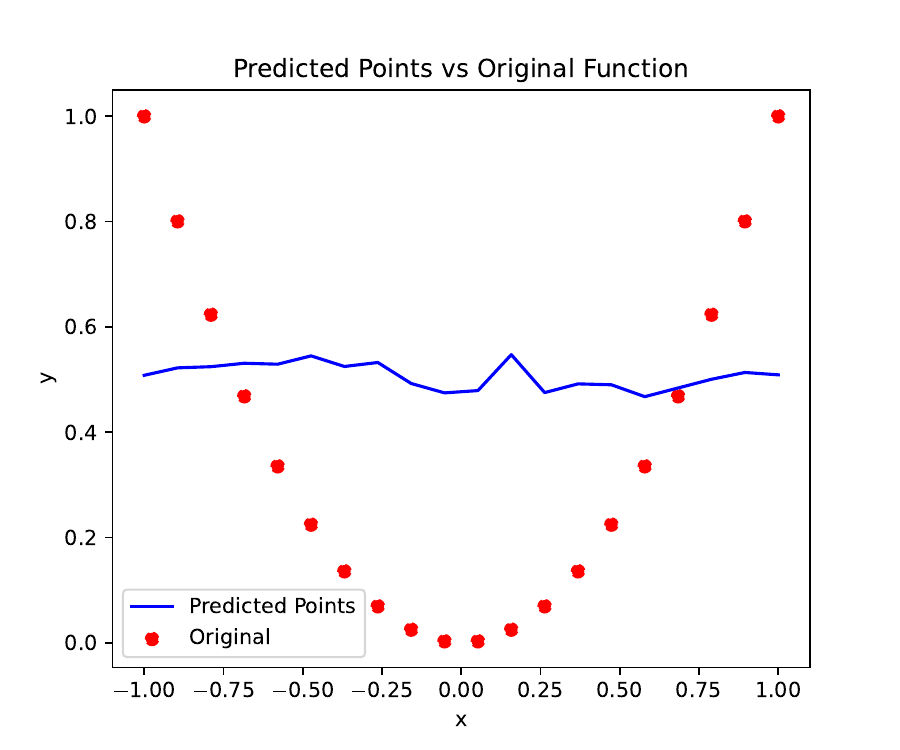}
    \caption{Result after using forward forward algorithm on physical neural networks with input layer containing trainable parameters.}
    \label{fig:pnnff_result}
\end{figure}

The results for the BP and FF-based regression for the simple function $y=x^2$ can be seen in figures \ref{fig:pnnbp_result} and \ref{fig:pnnff_result}, respectively. While the BP-based regression for DPNNs provide satisfactory convergence after around 15000 epochs, the FF-based DPNN provides no semblance of convergence after 10000 epochs of training. This indicates that further studies would be required to extend the FF-regression algorithms to DPNNs effectively.
    
\end{appendices}

\end{document}